\documentclass[lettersize,journal]{IEEEtran}
\usepackage{amsmath,amsfonts}
\usepackage{algorithmic}
\usepackage{algorithm}
\usepackage{array}
\usepackage{textcomp}
\usepackage{stfloats}
\usepackage{url}
\usepackage{verbatim}
\usepackage{graphicx}
\usepackage{subcaption}
\usepackage{cite}
\usepackage{tabularx}
\usepackage{multirow}
\usepackage{booktabs}
\usepackage{xcolor}
\usepackage{caption}
\usepackage[table]{xcolor}
\usepackage[colorlinks=true,linkcolor=black,citecolor=black,urlcolor=black]{hyperref}

\AtBeginDocument{%
  \fontdimen2\font=0.22em 
  \fontdimen3\font=0.1em  
  \fontdimen4\font=0.06em  
}

\begin{document}

\title{Reliable Microservice Tail Latency Prediction via Decoupled Dual-Stream Learning and Gradient Modulation}

\author{Wenzhuo Qian, Hailiang Zhao,~\IEEEmembership{Member,~IEEE,} Jiayi Chen, Ziqi Wang,~\IEEEmembership{Student Member,~IEEE,} Tianlv Chen, Zhiwei Ling, Xinkui Zhao, Kingsum Chow, Albert~Y.~Zomaya,~\IEEEmembership{Fellow,~IEEE,}, and Shuiguang Deng,~\IEEEmembership{Senior Member,~IEEE}

\thanks{Hailiang Zhao and Shuiguang Deng are the corresponding author.}
\thanks{Wenzhuo Qian and Shuiguang Deng are with College of Computer Science and Technology, Zhejiang University, Hangzhou 310027, China (e-mails: \{qwz, dengsg\}@zju.edu.cn).}
\thanks{Hailiang Zhao, Jiayi Chen, Ziqi Wang, Zhiwei Ling, Xinkui Zhao, Kingsum Chow are with School of Software Technology, Zhejiang University, Ningbo 315048, China (e-mails: \{hliangzhao, jyichen, wangziqi0312, zwling, zhaoxinkui, kingsum.chow\}@zju.edu.cn).}
\thanks{Tianlv Chen is with Polytechnic Institute, Zhejiang University, Hangzhou 310015, China (e-mails: \{22360327\}@zju.edu.cn).}
\thanks{Albert Y. Zomaya is with School of Computer Science, University of Sydney, Sydney, NSW 2006, Australia (e-mail: albert.zomaya@sydney.edu.au).}
}


\markboth{Journal of \LaTeX\ Class Files,~Vol.~14, No.~8, August~2021}%
{Shell \MakeLowercase{\textit{et al.}}: A Sample Article Using IEEEtran.cls for IEEE Journals} 


\maketitle

\begin{abstract}
Microservice architectures enable scalable cloud-native applications; however, the distributed nature of these systems complicates the maintenance of strict Service Level Objectives. Accurately predicting window-level P95 tail latency remains difficult due to the complex interactions between software workload propagation and infrastructure resource limits. Existing predictive models struggle to capture these dynamics because the lack of explicit separation between traffic metrics and resource metrics causes misaligned feature representations. Building on this suboptimal data treatment, the unified architectures of prior approaches fail to isolate cascading service dependencies from localized processing capacity. Due to this entanglement, joint training suffers from an optimization imbalance wherein resource features converge faster and dominate gradient updates, thereby preventing the learning of underlying software topologies. To address these challenges, we propose USRFNet, a dual-stream framework that separates the modeling of demand and capacity. The proposed framework utilizes a Graph Neural Network to model the spatial interactions of traffic workloads across software-level service dependencies, and a gating MLP to independently extract infrastructure-level resource dynamics. The model then integrates these representations through hierarchical tensor fusion. To resolve the training imbalance, we introduce a Reliability-Aware Gradient Modulation strategy that dynamically rescales gradients based on the generalization ratio of each data stream. Experiments on three large-scale real-world benchmarks demonstrate that USRFNet outperforms state-of-the-art methods in prediction accuracy. Specifically, compared to the best-performing baselines, the proposed framework achieves relative MAPE reductions ranging from $15.62\%$ to $26.11\%$ across the evaluated datasets.
\end{abstract}


\section{Introduction}

\begin{figure}[ht]
\centering
\includegraphics[width=0.95\columnwidth]{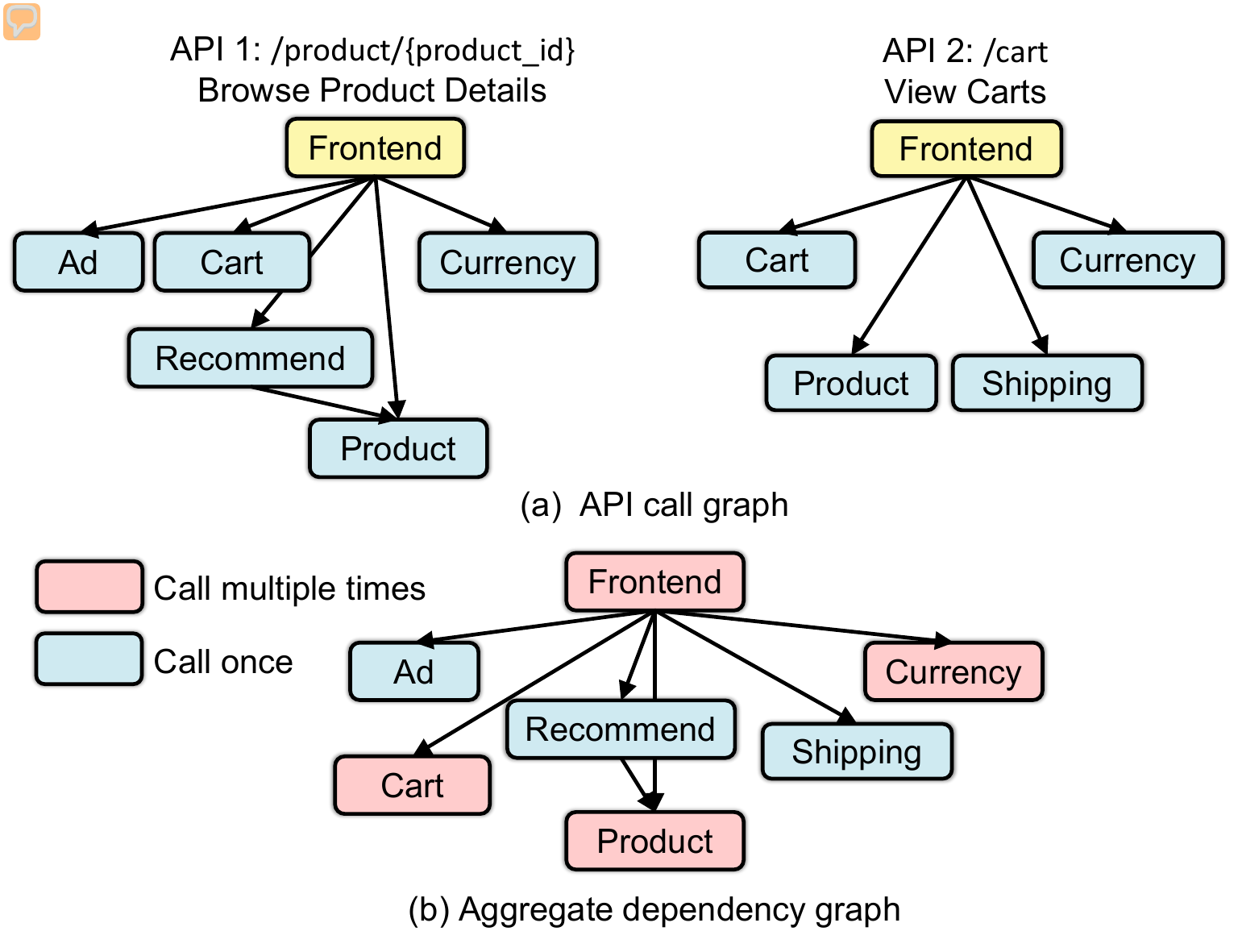}
\caption{Illustration of (a) an API call graph and (b) an aggregate dependency graph where multiple API calls are merged. The example is derived from the Online Boutique benchmark~\cite{onlineboutique2025}.}
\label{APIstate}
\end{figure}

\IEEEPARstart{I}n recent years, microservice architecture has emerged as the standard for building scalable cloud-native applications~\cite{dragoni2017microservices,sriraman2019softsku}. By decomposing monolithic applications into fine-grained and loosely coupled components, this paradigm enables independent deployment and operational agility, as illustrated in Fig. \ref{APIstate}. However, this architectural decomposition complicates system reliability assurance; while the overall system can be conceptualized as an aggregate dependency graph (Fig.~\ref{APIstate}b), the distributed nature transforms simple function calls into complex remote procedure calls across a dynamic topology. This transformation creates substantial challenges in maintaining strict Service Level Objectives. In such environments, performance degradation propagates across complex dependency chains~\cite{liu2021microhecl}, which transforms proactive resource allocation and capacity planning into a high-dimensional optimization problem.

Proactive reliability management requires anticipating system performance. Although existing studies predominantly focus on predicting the latency of individual requests~\cite{tam2023pert, xu2025fastpert}, this fine-grained granularity is often insufficient for system-level decisions. The latency of individual requests is highly stochastic and sensitive to transient noise, such as garbage collection pauses or network jitter, which may not reflect actual service degradation. In contrast, forecasting window-level P95 latency provides a stable metric for evaluating the distribution of tail latency. Recent advancements~\cite{park2021graf} indicate that proactive resource allocation primarily requires estimating the steady-state equilibrium between the current system demand and the target infrastructure capacity. By adopting a Markovian approximation of system states, tail latency spikes within a micro-scale time window are considered to arise from this immediate mismatch rather than from the slow accumulation of historical data. Therefore, formulating the prediction as a snapshot spatial regression and deliberately discarding historical dependencies effectively strips away long-term time-series noise, thereby revealing overall system trends and critical performance bottlenecks. Such forecasting serves as a foundation for macro-level tasks, including automated horizontal scaling and vertical resource orchestration. However, accurate prediction remains difficult because of the complex nature of microservice observability data. Specifically, we identify three key challenges that limit the effectiveness of existing predictors:

\begin{enumerate} 
    \item \textit{Challenge 1: Representation Misalignment of Heterogeneous Metrics.} Microservice performance results from the interaction between global traffic propagation dynamics and local resource saturation constraints~\cite{somashekar2024gamma, sun2024art}. Existing approaches~\cite{park2021graf, luo2022erms,tam2023pert, xu2025fastpert} often process these distinct signals through unified encoding schemes and disregard metric heterogeneity. This misalignment prevents the effective modeling of causal relationships between specific metrics and system latency, which potentially masks early warning signs of performance degradation.

    \item \textit{Challenge 2: Architectural Entanglement of Distinct Mechanics.} Current predictors~\cite{park2021graf, luo2022erms,tam2023pert, xu2025fastpert} often fuse heterogeneous data streams into a shared feature space without recognizing their fundamentally different physical origins. In cloud-native environments, traffic paths can be explicitly traced via network protocols to form explicit topologies. Conversely, the coexistence of and resource contention among microservice instances on underlying host machines remain implicit and invisible because of the black-box scheduling mechanisms of cloud providers. This entanglement fails to address the structural dichotomy between explicit traffic propagation and implicit infrastructure interference, leading to inaccurate representations of the system state.

    \item \textit{Challenge 3: Optimization Imbalance via Convergence Disparity.} The joint training of heterogeneous components introduces a significant optimization discrepancy~\cite{peng2022balanced}. Metrics on the resource stream, which correlate directly with performance targets, typically generate larger gradients and converge more rapidly than traffic stream features that involve high-order topological dependencies. This convergence disparity biases the optimization trajectory towards the resource stream and suppresses the gradient updates of the components responsible for traffic dynamics.
\end{enumerate}

To solve these issues, we propose USRFNet, a dual-stream framework designed for the reliable prediction of window-level P95 latency. To resolve the representation misalignment and architectural entanglement, USRFNet uses a physically decoupled encoding architecture. A topology-aware Graph Neural Network~\cite{scarselli2008graph, shi2021masked} explicitly models the propagation of system demand across service dependencies, while a gated Multilayer Perceptron network~\cite{liu2021pay} independently encodes the dynamics of infrastructure capacity. Furthermore, to mitigate the optimization imbalance inherent in joint training, we design a Reliability-Aware Gradient Modulation mechanism. This component dynamically monitors the generalization ratio between the topological learning of traffic metrics and the rapid convergence of resource metrics. It explicitly rescales the gradients of each stream to prevent the dominant stream from controlling the training process. Finally, these physically distinct representations are synthesized through a hierarchical fusion module to generate comprehensive system embeddings.

The primary contributions of this paper are summarized as follows:

\begin{itemize}
     \item We propose USRFNet, a physically decoupled dual-stream framework that aligns the architecture of the model with the physical mechanics of microservice performance. By explicitly separating the modeling of cascading traffic demand from localized infrastructure capacity, the framework effectively addresses the semantic gap in heterogeneous observability data and overcomes the limitations of traditional flat fusion strategies.

     \item We design the Reliability-Aware Gradient Modulation mechanism to address the convergence heterogeneity between traffic and resource data streams. This mechanism resolves the inherent disparity in learning rates by rescaling gradients based on dynamic generalization ratios. This adjustment allows the model to fully use the feature characteristics from both data streams for enhanced predictive performance.

     \item We conduct experiments on large-scale datasets collected from three real-world microservice benchmarks: Online Boutique~\cite{onlineboutique2025}, Sock Shop~\cite{sockshop2025}, and Train Ticket~\cite{zhou2018benchmarking}. Furthermore, the datasets collected in this study have been open-sourced to support future research in software performance engineering\footnote{\texttt{https://zenodo.org/records/18728725}}. The evaluation results show that USRFNet significantly outperforms advanced baselines in prediction accuracy and exhibits strong reliability. This strong performance provides a reliable foundation for downstream tasks, such as proactive resource orchestration and autoscaling.
\end{itemize}

\section{Motivation}

This section motivates our study by addressing three key questions: (1) why window-level P95 latency is the prediction target; (2) why heterogeneous demand and capacity signals must be decoupled; and (3) why gradient modulation is necessary for balanced joint training. The first question explores an optimal performance metric to support reliable system-level decisions. The second question justifies the necessity of a dual-stream architecture for modeling heterogeneous observability data and mitigating signal interference. The third question analyzes the optimization discrepancies that emerge during joint training, which motivates the design of dynamic gradient modulation for balanced learning.

\subsection{Why Window-Level P95 Latency as the Prediction Target?}
\begin{figure}[ht]
\centering
\includegraphics[width=0.7\columnwidth]{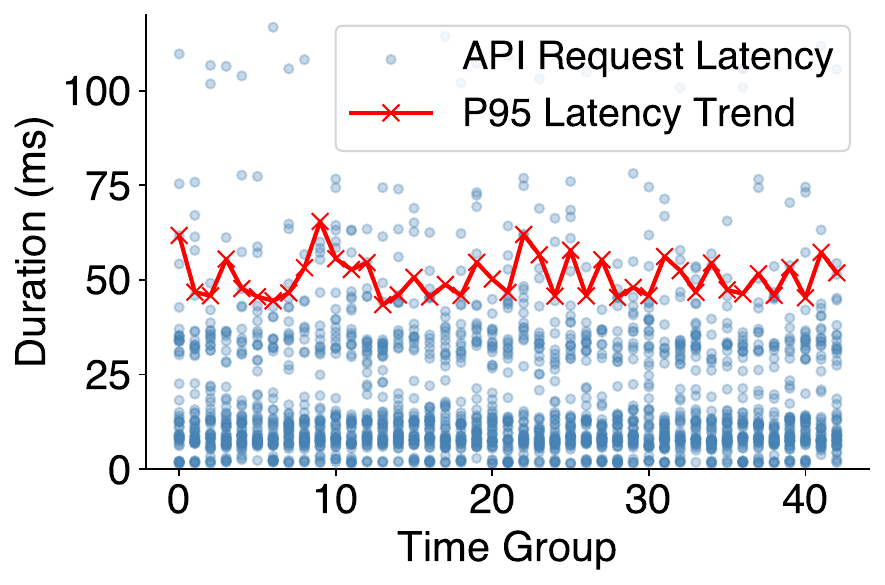} 
\caption{Volatility comparison between raw request latencies (blue dots) and smoothed window-level P95 latency (red line). Individual request latencies are highly variable and sensitive to transient noise, whereas aggregation over 5-second intervals yields a stable metric reflecting overall system performance trends.}
\label{motivation1}
\end{figure}

The prediction of microservice performance requires a stable and representative metric. While the latency of individual API requests is intuitive, it is highly volatile and sensitive to transient noise, as demonstrated by the scattered blue data points in Fig.~\ref{motivation1}. Reacting to these stochastic fluctuations can introduce instability into automated decision-making processes and trigger unnecessary scaling operations. Therefore, we target the prediction of window-level P95 latency, which is defined as the 95th percentile over a fixed time window. As illustrated by the smoothed red line in Fig.~\ref{motivation1}, this aggregation provides a significantly more stable target. Aggregating latencies explicitly captures the tail-end user experience, which serves as a sensitive indicator of system degradation. Furthermore, this aggregation smooths out transient noise and reveals underlying performance trends that represent actual changes in the system state. Such stability is essential for system-level operations, such as proactive resource management and anomaly detection.

\subsection{Why Decouple Heterogeneous Demand and Capacity Signals?}
The effective modeling of microservice demand and capacity requires an explicit recognition of their distinct characteristics and interactions~\cite{somashekar2024gamma, sun2024art}. The performance of a microservice depends on the interaction between system demand, which is primarily represented by traffic metrics, and infrastructure capacity, which is primarily characterized by resource metrics. These two aspects operate under fundamentally different mechanisms. Traffic metrics, such as request rate and throughput, reflect event-driven workload dynamics and directly measure the load placed on the system. Conversely, resource metrics, such as the utilization of CPU and memory, represent the processing capability of the system and evolve with different temporal patterns. Fusing these physically distinct signals into a single processing stream creates semantic misalignment, which causes feature interference and suboptimal representations.

\begin{figure}[ht]
\centering
\includegraphics[width=0.7\columnwidth]{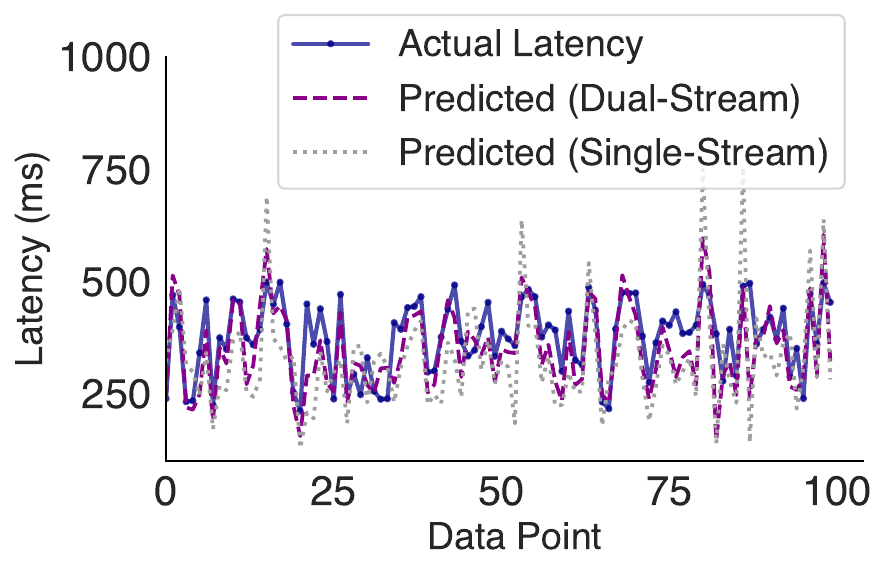} 
\caption{Performance comparison of single-stream and dual-stream architectures. The single-stream GNN produces volatile predictions with large deviations, whereas the decoupled dual-stream model closely tracks the ground truth, confirming the need to separate heterogeneous signals.}
\label{motivation2}
\end{figure}

To empirically assess this impact, we compared a single-stream Graph Neural Network with a dual-stream prototype. As illustrated in Fig.~\ref{motivation2}, the single-stream model produced volatile predictions that deviated significantly from the ground truth. The mismatched dynamics between traffic metrics (which represent demand causes) and resource metrics (which reflect capacity effects) induced severe signal interference when processed jointly. This observation motivates the dual-stream design of USRFNet, which explicitly separates the representation learning of these heterogeneous streams. By independently encoding traffic features and resource features, the model prevents signal interference and captures their distinct causal roles prior to integration. This architectural decoupling is necessary to achieve an accurate understanding of the state of the microservice system.

\subsection{Why Gradient Modulation for Balanced Joint Training?}
\label{balanced_learning}
\begin{figure}[ht]
\centering
\includegraphics[width=0.7\columnwidth]{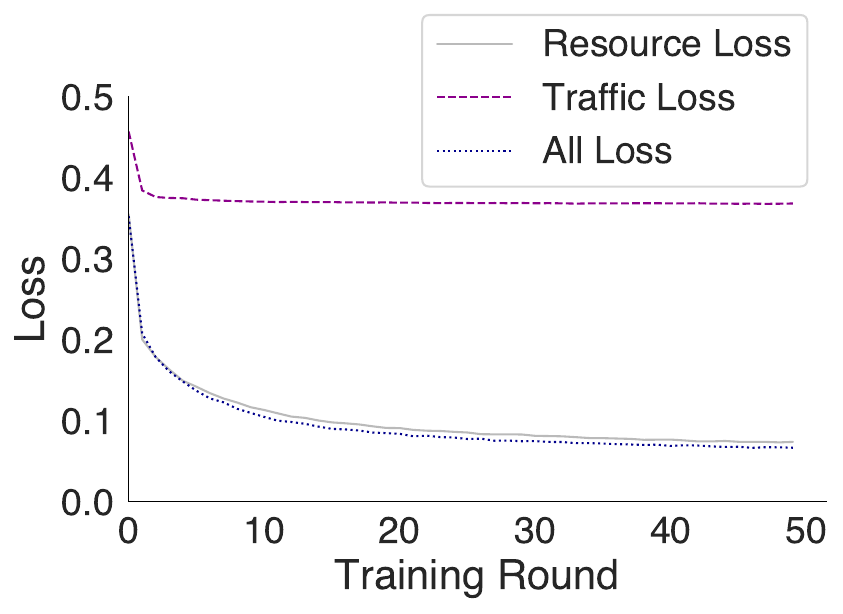}
\caption{Training loss trajectories showing optimization interference. The rapidly converging resource stream (gray) dominates the training process, suppressing the traffic stream (purple). This imbalance causes structural bias and necessitates gradient modulation.}
\label{motivation3}
\end{figure}

The joint training of heterogeneous data streams frequently introduces an optimization imbalance. As depicted in Fig.~\ref{motivation3}, the resource stream, represented by the gray line, acts as the dominant branch by converging rapidly and stabilizing early. This phenomenon occurs because resource metrics correlate directly with system latency, which provides more immediate feedback to the optimizer compared to the complex topological dependencies encapsulated within the traffic stream. Conversely, the traffic stream, represented by the purple line, exhibits sluggish optimization dynamics and maintains a higher loss throughout the training phase.

This convergence discrepancy induces detrimental signal interference. The rapidly converging resource branch generates gradients with significantly larger magnitudes~\cite{peng2022balanced, tao2024giving, huang2025adaptive}. These gradients dominate the optimization process and suppress the parameter updates of the lagging traffic branch. Consequently, the shared optimization trajectory becomes biased. The model overfits to the fast-converging infrastructure capacity signals and neglects the structural insights embedded within the service dependency graph. This structural bias prevents the model from capturing performance degradations caused by interaction anomalies that are not reflected in resource saturation metrics.

A naive loss summation cannot resolve this dynamic imbalance, which motivates the design of a dedicated gradient modulation mechanism described in Section~\ref{sec:ragm}.

\section{Preliminary}

\begin{figure*}[!ht]
\centering
\includegraphics[width=0.8\textwidth]{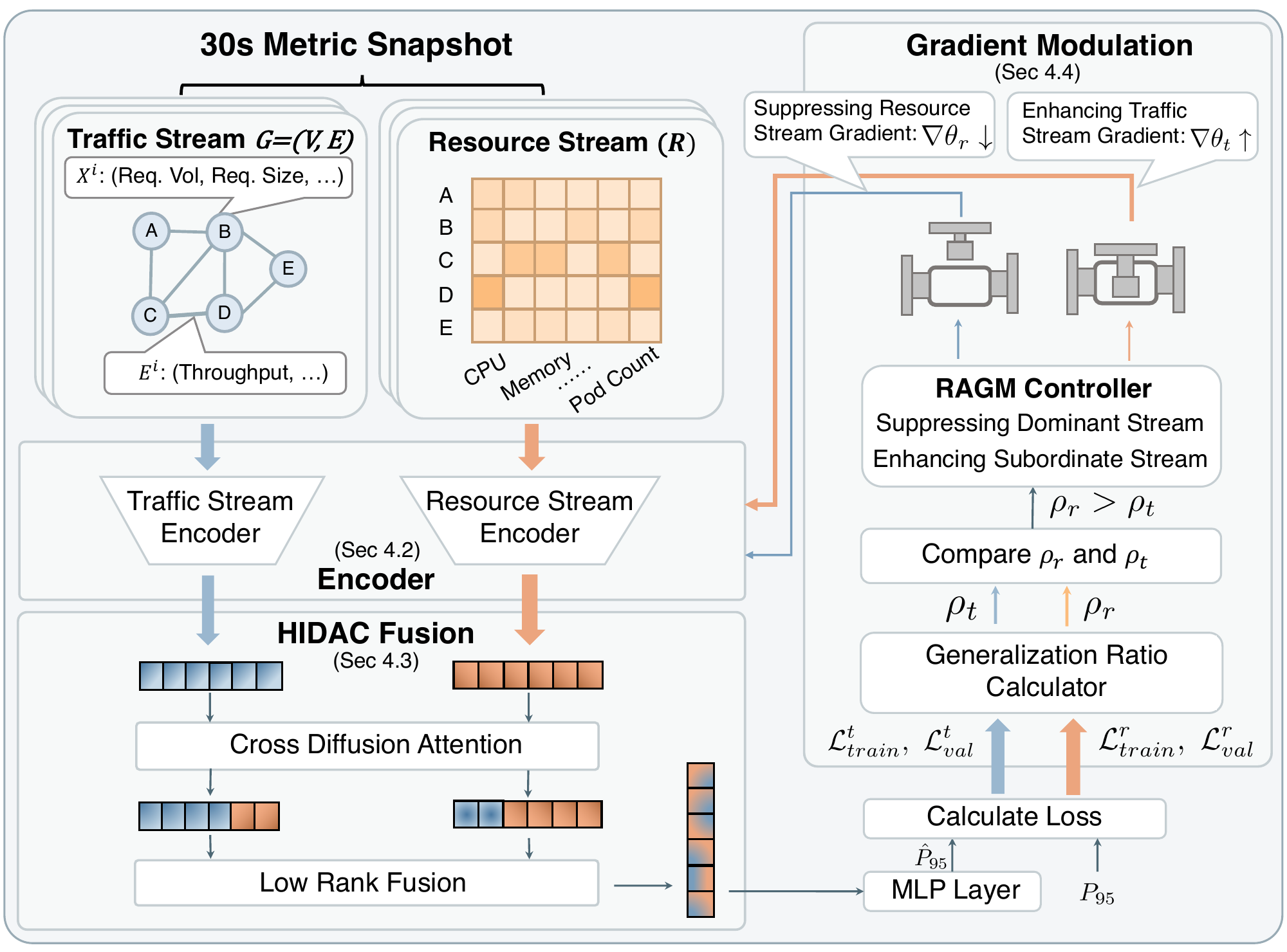} 
\caption{Architecture of USRFNet. The model employs two physically decoupled encoders to extract the features of traffic propagation and resource saturation. The HIDAC module hierarchically fuses these representations through context alignment and tensor fusion, while the RAGM controller dynamically scales the gradients to ensure balanced training.}
\label{figure1}
\end{figure*}

\subsection{Optimization Imbalance in Heterogeneous Learning}
\label{sec:prelim_imbalance}
Jointly training on heterogeneous data streams frequently suffers from optimization imbalance~\cite{peng2022balanced, tao2024giving, huang2025adaptive}: streams with simpler patterns converge faster, generating larger gradients that suppress updates of lagging streams and produce biased representations. This well-known disparity provides the theoretical foundation for the gradient regulation strategy described in Section~\ref{sec:ragm}.

\subsection{Neural Networks}
This section introduces the neural architectures used to capture topological propagation and resource dynamics.

\paragraph{Transformer-based Graph Convolution.} Traditional GCNs~\cite{kipf2017semisupervised} aggregate features by using static spectral filters, which assigns fixed importance based on structural degrees. This rigid approach cannot capture the fluctuating intensity of microservice traffic, in which low-degree nodes often handle critical high-throughput loads. While GATs~\cite{veličković2018graph} introduce attention mechanisms, they typically compute importance scores based solely on node feature similarity, thereby ignoring the explicit interaction attributes, such as throughput or latency, carried by the edges. To solve this, we use the Transformer-based Graph Convolution~\cite{shi2021masked}, which extends the message-passing framework with multi-head self-attention that explicitly includes edge features. This modification allows the model to dynamically weight the neighboring nodes based on the actual traffic volume rather than solely on semantic similarity. We adopt this architecture to model the dynamic nature of system demand, which ensures that the model prioritizes critical high-volume dependencies over background traffic.

\paragraph{Gated Multilayer Perceptron (gMLP).} Recent studies in tabular deep learning, such as FT-Transformer~\cite{gorishniy2021revisiting}, show that applying Transformer architectures to feature embeddings captures complex cross-feature dependencies via self-attention~\cite{vaswani2017attention}. In cloud-native systems, however, the evaluation of infrastructure capacity requires modeling the implicit physical resource competition among services. Standard Graph Neural Networks require explicit edges, which are unavailable for unobservable host-level contention. While Transformers capture dense dependencies, they lack the specific mechanisms required to separate spatial hardware interference from general feature interactions. The gMLP model~\cite{liu2021pay} addresses this issue through a pure MLP-based architecture that uses a Spatial Gating Unit (SGU). Instead of relying on predefined edges, the SGU uses learnable linear projections across the spatial dimension to explicitly capture the global dependencies among all input nodes. Therefore, USRFNet uses the gMLP model as the resource stream encoder to overcome the limitations of explicit topologies. This component uses the spatial gating mechanism to infer unobservable correlations and implicit contention patterns from the infrastructure metrics.

\subsection{Problem Formulation}
Our objective is to predict the P95 end-to-end latency over a short, fixed-length time window. We set the window length to 30 seconds, which matches common auto-scaling intervals and provides sufficient statistical stability for timely operational decisions.

We define the structural bounds of the application as an aggregate dependency graph $\mathcal{G} = (\mathcal{V}, \mathcal{E})$, where $\mathcal{V}$ and $\mathcal{E}$ represent the fixed sets of microservices and potential call dependencies, respectively. To accurately reflect cloud-native environments, we distinguish between static service-level dependencies and highly dynamic infrastructure-level capacities. The aggregate graph $\mathcal{G}$ provides a fixed structural backbone that represents the stable business logic (i.e., potential API call paths), but the actual system experiences severe volatility at the infrastructure level because of automated pod scaling and resource redistribution. Instead of forcing an unstable graph structure to model these micro-level fluctuations, USRFNet maintains the stable service-level topology $\mathcal{G}$ and captures the infrastructure dynamics through time-variant node, edge, and resource feature matrices. The training dataset
\begin{equation}
    \mathcal{D} = \left\{\mathbf{X}^{(i)}, \mathbf{E}^{(i)}, \mathbf{R}^{(i)}, y^{(i)}\right\}_{i=1}^{T}, 
\end{equation}
consists of $T$ snapshots, which are indexed by the observation window $(i)$.

The node feature matrix $\mathbf{X}^{(i)} \in \mathbb{R}^{|\mathcal{V}| \times d_n}$ captures the traffic stream features for each microservice, such as the send and receive throughput. The edge feature matrix $\mathbf{E}^{(i)} \in \mathbb{R}^{|\mathcal{E}| \times d_e}$ encodes the traffic stream interactions between services, such as the request and response throughput. Additionally, the matrix $\mathbf{R}^{(i)} \in \mathbb{R}^{|\mathcal{V}| \times d_r}$ contains the resource stream features, such as the utilization of CPU and memory, and the pod counts. A comprehensive description of these selected dataset features is provided in Appendix~A1. Finally, the scalar $y^{(i)} \in \mathbb{R}$ denotes the ground-truth P95 end-to-end latency observed during the $i$-th window. Here, $d_n$, $d_e$, and $d_r$ denote the feature dimensions of the traffic stream node features, the traffic stream edge features, and the resource stream features, respectively. In the subsequent equations, we omit the superscript ``$(i)$'' for simplicity.

We formulate the prediction of window-level latency as a supervised graph regression task. The goal is to learn a function
\begin{equation}
    \hat{y} = f_{\theta} \big(\mathcal{G}, \mathbf{X}, \mathbf{E}, \mathbf{R}\big),
\end{equation}
that maps the input system features to the predicted P95 end-to-end latency. Here, $f_{\theta}$ represents the prediction model, and $\hat{y}$ is the target prediction.

\section{Design}

\subsection{USRFNet Overview}
To model heterogeneous system dynamics and resolve optimization imbalances, we propose USRFNet (\underline{U}nified \underline{S}ystem \underline{R}epresentation \underline{F}usion \underline{Net}work). As illustrated in Fig. \ref{figure1}, the framework uses a physically decoupled dual-stream architecture that processes traffic metrics and resource metrics through three synergistic components: decoupled representation learning, hierarchical integration, and reliability-aware gradient modulation. 

The workflow begins with the decoupled representation learning phase. Parallel encoders independently model the topological propagation of system demand using transformer-based graph convolutions~\cite{shi2021masked} and the saturation dynamics of infrastructure capacity using gated MLPs~\cite{liu2021pay}. The Hierarchical Integration of Demand And Capacity (HIDAC) module then bridges the semantic gap by fusing these representations into a comprehensive system embedding for predicting window-level P95 latency. During training, the Reliability-Aware Gradient Modulation (RAGM) mechanism dynamically monitors generalization discrepancies and scales the gradients. This mechanism synchronizes convergence between the heterogeneous streams, which ensures stable structural generalization and prevents the optimization trajectory from becoming biased towards simpler features.

\subsection{Decoupled Representation Learning}
\label{sec:decoupled_learning}
\begin{figure}[ht]
    \centering
    \begin{subfigure}[b]{0.9\columnwidth}
        \centering
        \includegraphics[width=\linewidth]{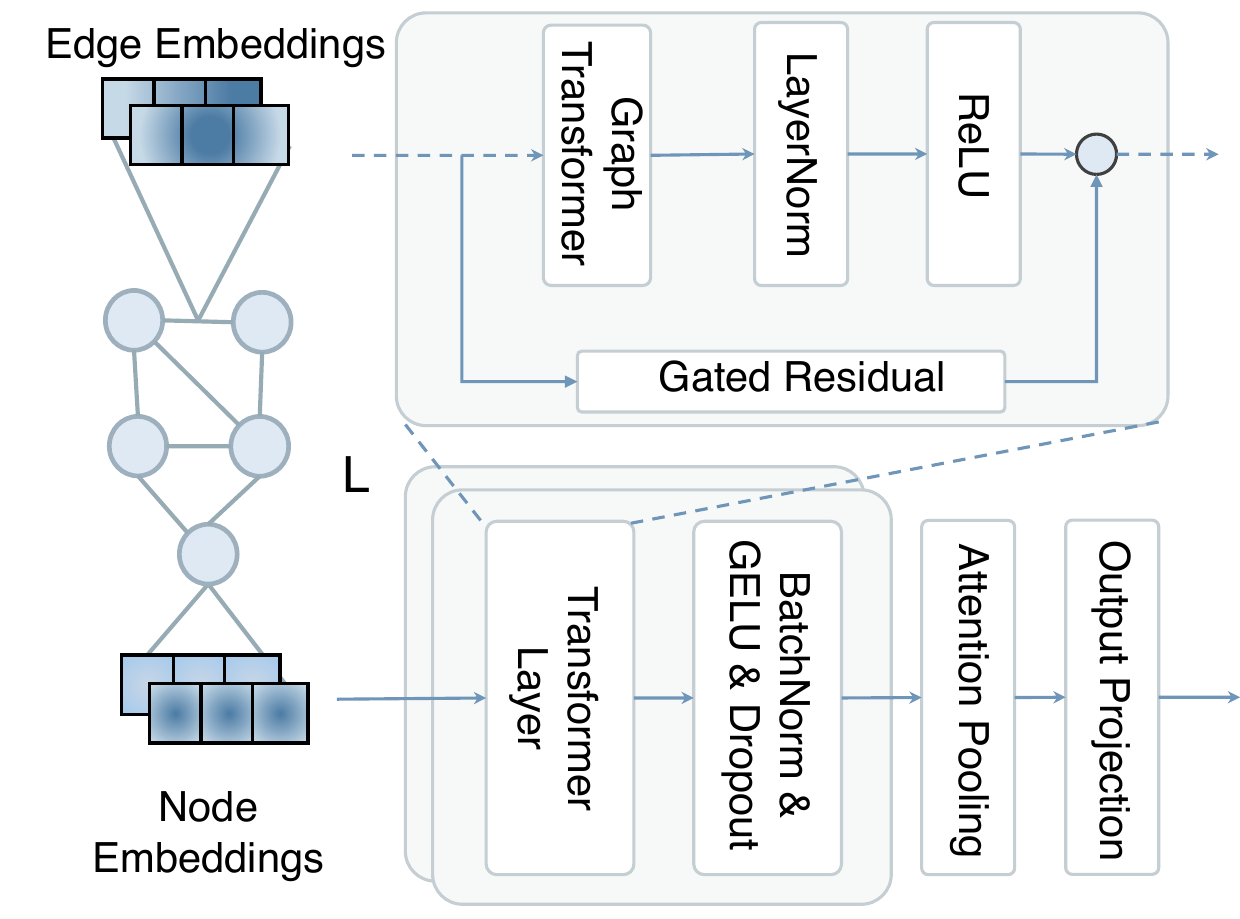} 
        \caption{Traffic Stream Encoder}
        \label{fig:traffic_enc}
    \end{subfigure}
    
    \vspace{10pt} 

    \begin{subfigure}[b]{0.95\columnwidth}
        \centering
        \includegraphics[width=\linewidth]{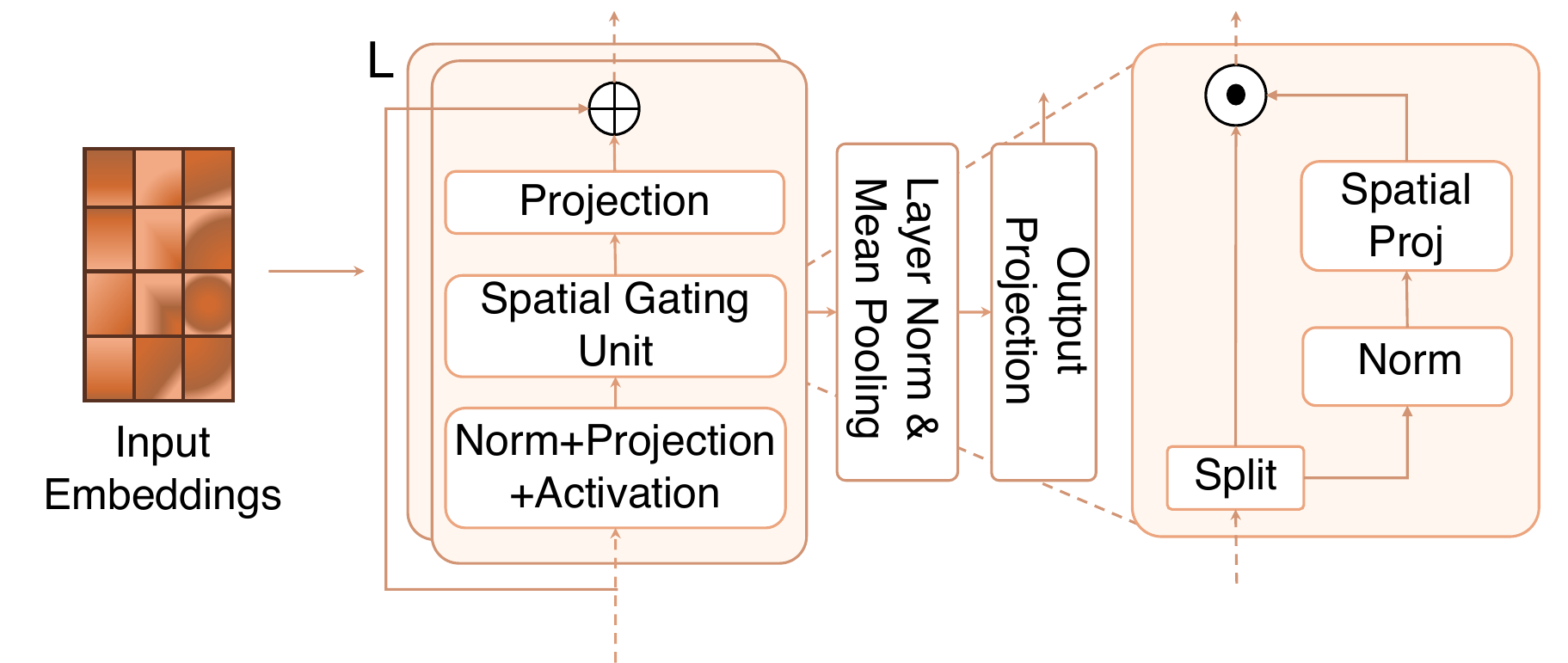}
        \caption{Resource Stream Encoder}
        \label{fig:resource_enc}
    \end{subfigure}

    \caption{Detailed architecture of the Decoupled Representation Learning module. (a) The traffic stream Encoder uses Transformer-based Graph Convolution to aggregate demand features based on edge intensity. (b) The resource stream Encoder uses a Gated MLP with a split-head mechanism to capture global capacity correlations.}
    \label{fig:encoders}
\end{figure}

To resolve representation misalignment and architectural entanglement, we design two specialized encoders, as depicted in Fig. \ref{fig:encoders}. This decoupled architecture ensures that propagation-based traffic signals and saturation-based resource signals are modeled by mechanisms that match their physical characteristics.

\subsubsection{Modeling System Demand via traffic stream Encoder}
As illustrated in Fig. \ref{fig:traffic_enc}, the traffic encoding process begins by formalizing the aggregate dependency graph $\mathcal{G}=(\mathcal{V}, \mathcal{E})$ as the structural backbone. We construct node features $\mathbf{X} \in \mathbb{R}^{N \times d_n}$ and edge features $\mathbf{E} \in \mathbb{R}^{|\mathcal{E}| \times d_e}$ from workload statistics, such as request volume, request size, and throughput. Here, node features represent the localized workload accumulation, while edge features quantify the traffic volume flowing between services.

The core processing unit consists of stacked transformer-based graph convolution layers~\cite{shi2021masked}. To integrate invocation intensity into feature extraction, the model uses a multi-head attention mechanism operating within a projected subspace. For a specific head $k$, we first map the node representations $h$ and edge attributes $e_{uv}$ into a shared latent space of dimension $d_k$ using learnable projection matrices $W_Q^{(k)}, W_K^{(k)}, W_E^{(k)}$. The edge-integrated attention coefficient $\alpha_{uv}^{(k)}$ is then computed as:
\begin{equation}
    \alpha_{uv}^{(k)} = \text{softmax} \left( \frac{(h_u W_Q^{(k)})(h_v W_K^{(k)} + e_{uv} W_E^{(k)})^T}{\sqrt{d_k}} \right).
\end{equation}
Here, $\sqrt{d_k}$ serves as a scaling factor to stabilize the gradients. Physically, the term $(h_v W_K^{(k)} + e_{uv} W_E^{(k)})$ represents a traffic-modulated key, where the static features of the upstream neighbor $v$ are dynamically enhanced by the corresponding emitted traffic volume $e_{uv}$. High-throughput edges amplify the semantic alignment with the query node $u$, which leads the encoder to assign larger aggregation weights to critical dependencies. This mechanism transforms the static backbone into a traffic-driven dynamic topology. When an invocation path becomes inactive because of instance failures or routing shifts, the corresponding edge feature $e_{uv}$ naturally decays. This dynamically masks the attention weight, effectively performing dynamic topology pruning at each time window. The updated node representations are aggregated via an attention-based pooling layer, producing a demand-aware embedding $\mathbf{z}_t$ that reflects the global workload distribution.

\subsubsection{Modeling Infrastructure Capacity via resource stream Encoder}
Parallel to the traffic stream, Fig. \ref{fig:resource_enc} illustrates the workflow of the resource encoder. Resource metrics, such as the utilization of CPU and memory, form a sequence of service features $\mathbf{R} \in \mathbb{R}^{N \times d_r}$. The gMLP architecture~\cite{liu2021pay} processes this sequence to model global dependencies through static spatial projections. The core component of this encoder is the Spatial Gating Unit (SGU), which applies a three-stage transformation to capture capacity bottlenecks. 

The encoding process begins by projecting the input features $\mathbf{R}$ into a higher-dimensional latent space $\mathbf{Z} \in \mathbb{R}^{N \times d}$ using a linear expansion layer, which provides representational capacity for subsequent transformations. For efficient interaction, the model splits $\mathbf{Z}$ into two independent halves, $\mathbf{Z}_1$ and $\mathbf{Z}_2$, along the channel dimension. A spatial projection is applied to $\mathbf{Z}_2$ to capture the global context:
\begin{equation}
    f_{W}(\mathbf{Z}_2) = W_{spatial} \cdot \mathbf{Z}_2 + \mathbf{b}.
\end{equation}
The learnable matrix $W_{spatial} \in \mathbb{R}^{N \times N}$ operates over the sequence dimension (the set of services) rather than the feature channel dimension. Because the black-box scheduling of cloud providers prevents the observation of the true service-host bipartite graph, resource features cannot rely on explicit graph convolutions. Instead, this sequence-wise matrix multiplication allows the resource state of every service to interact linearly with all other services. This operation constructs a fully connected virtual competition graph in the latent space, which allows the model to infer unobservable physical hardware interference through representation learning.

The spatially enhanced features are then fused with the original signals using a multiplicative gating operation:
\begin{equation}
    \mathbf{Z}' = \mathbf{Z}_1 \odot f_{W}(\mathbf{Z}_2),
\end{equation}
where $\odot$ denotes element-wise multiplication. Through this mechanism, the encoder captures global dependencies without constructing an explicit topology or relying on predefined edges. The processed features are then compressed using global average pooling to generate the capacity-aware embedding $\mathbf{z}_r$.

\subsection{Hierarchical Integration of Demand and Capacity (HIDAC)}
\label{sec:hidac}
\begin{figure}[ht]
    \centering
    \includegraphics[width=0.9\columnwidth]{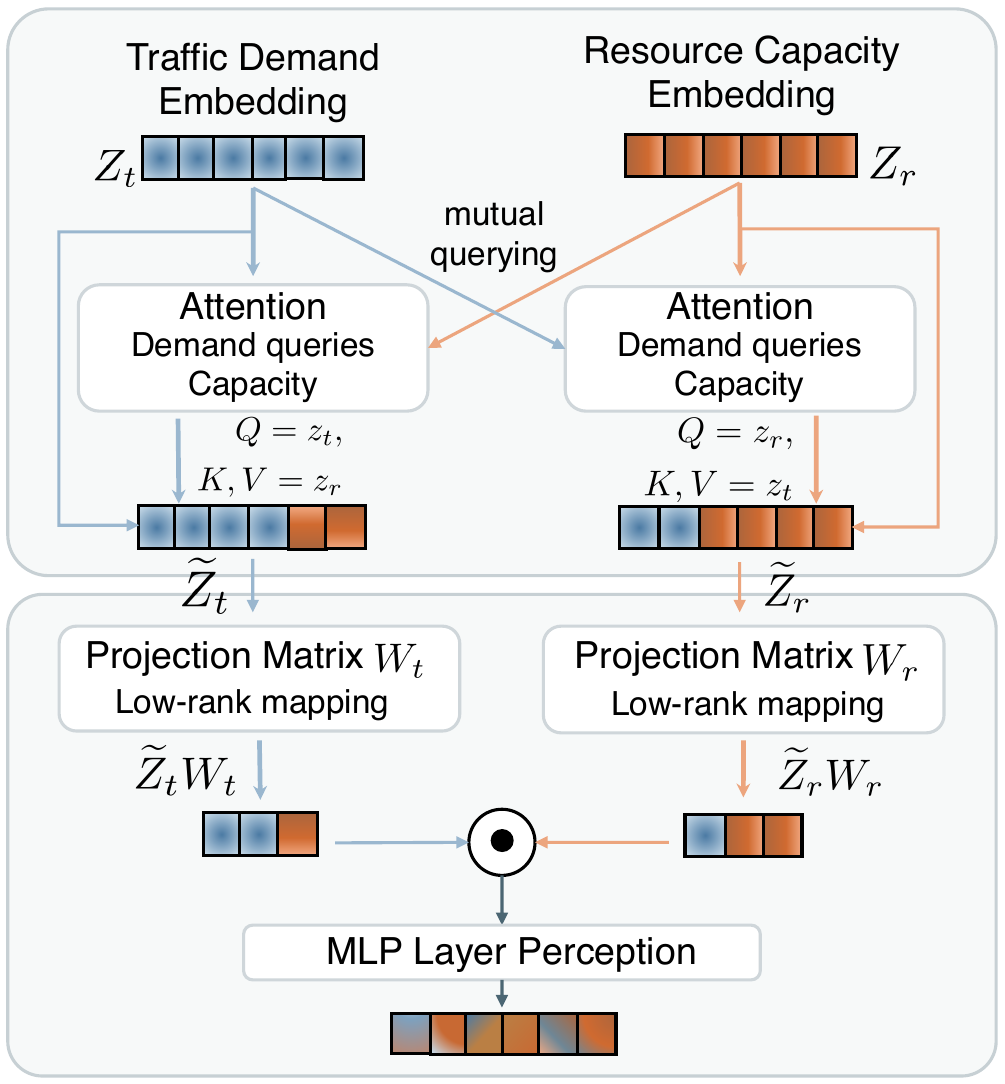} 
    \caption{Workflow of the HIDAC module. The process consists of two phases: (1) Contextual Alignment, where Cross-Diffusion-Attention enables mutual querying between traffic and resource streams to identify bottlenecks; (2) Low-Rank Tensor Fusion, which models the multiplicative demand-capacity interactions via element-wise operations in a projected subspace.}
    \label{fig:hidac}
\end{figure}

Existing predictors often use naive concatenation to merge heterogeneous streams. This approach fails to capture the complex non-linear interactions between system demand and infrastructure capacity. To address this architectural limitation, we propose the HIDAC module. As illustrated in Fig. \ref{fig:hidac}, this module uses a hierarchical two-stage fusion strategy to combine the physically distinct embeddings $\mathbf{z}_t$ and $\mathbf{z}_r$ into a unified representation.

The first stage, contextual alignment, begins by aligning the distinct semantic spaces of the two modalities. We use Cross-Diffusion-Attention~\cite{wang2024mutualformer} to enable mutual querying between the traffic stream and the resource stream. As depicted in the upper part of Fig. \ref{fig:hidac}, this mechanism allows the traffic demand vector to query the resource capacity vector to identify potential bottlenecks ($Q=\mathbf{z}_t, K=V=\mathbf{z}_r$), while simultaneously allowing the resource vector to perceive the workload patterns that cause system saturation. To ensure numerical stability and preserve signal integrity during cross-modal injection, we apply a residual structure followed by a normalization layer. The enhanced embeddings $\tilde{\mathbf{z}}_t$ and $\tilde{\mathbf{z}}_r$ are computed as follows:
\begin{align}
    \tilde{\mathbf{z}}_t &= \text{Norm}(\mathbf{z}_t + \text{Attention}(Q=\mathbf{z}_t, K=\mathbf{z}_r, V=\mathbf{z}_r)), \\
    \tilde{\mathbf{z}}_r &= \text{Norm}(\mathbf{z}_r + \text{Attention}(Q=\mathbf{z}_r, K=\mathbf{z}_t, V=\mathbf{z}_t)),
\end{align}
where $\text{Attention}(\cdot)$ denotes the standard multi-head attention operation~\cite{vaswani2017attention} and $\text{Norm}(\cdot)$ represents layer normalization. This design ensures that the traffic representations are dynamically calibrated by infrastructure constraints within a stabilized feature space. This calibration facilitates the identification of demand-capacity misalignments.

The second stage, low-rank tensor fusion, captures the high-order correlations between the aligned features. The interaction between demand and capacity is inherently multiplicative (e.g., latency spikes exponentially when demand exceeds capacity). To explicitly model this dynamic, we use Low-Rank Tensor Fusion~\cite{liu2018efficient}, as illustrated in the lower part of Fig. \ref{fig:hidac}. Instead of computing the computationally expensive outer product, we project the embeddings into a shared subspace by using learnable matrices $W_t$ and $W_r$, which is followed by an element-wise product:
\begin{equation}
    \mathbf{z}_{f} = \phi \left( (\tilde{\mathbf{z}}_t W_t) \odot (\tilde{\mathbf{z}}_r W_r) \right),
\end{equation}
where $\odot$ denotes the Hadamard product, and $\phi$ is a multilayer perceptron that maps the fused representation to the final latent space. This operation efficiently approximates the tensor product to yield a unified system embedding $\mathbf{z}_{f}$. This embedding comprehensively integrates demand propagation and capacity constraints for the downstream latency predictor.

\subsection{Reliability-Aware Gradient Modulation (RAGM)}
\label{sec:ragm}
The joint training of heterogeneous streams introduces a significant optimization imbalance. Because resource stream metrics are statistically simpler and correlate more directly with performance targets, they typically converge faster than topologically complex traffic metrics. This disparity causes the resource stream to dominate the gradient updates, which effectively suppresses the learning process of the traffic encoder. To resolve this issue, we introduce the RAGM mechanism to dynamically synchronize the learning pace.

To assess generalization quality without causing data leakage from the final testing set, we adopt a generalization-guided evaluation strategy inspired by recent multimodal adaptive optimization frameworks~\cite{tao2024giving}. Specifically, we monitor the loss reduction on both the standard training batches and a held-out validation set at each training step. Let $\Delta \mathcal{L}_{train}^{k, (t)}$ and $\Delta \mathcal{L}_{val}^{k, (t)}$ denote the loss reduction for stream $k \in \{Traffic, Resource\}$ at epoch $t$. To ensure numerical stability and prevent anomalous gradient scaling during severe loss oscillations, we lower-bound the loss reductions with a small positive constant $\epsilon$ (i.e., $\Delta \mathcal{L} = \max(\Delta \mathcal{L}, \epsilon)$) prior to the computation. We define the generalization ratio $\rho_k^{(t)}$ as the efficiency of knowledge transfer:
\begin{equation}
    \rho_k^{(t)} = \frac{\Delta \mathcal{L}_{val}^{k, (t)}}{\Delta \mathcal{L}_{train}^{k, (t)}}.
\end{equation}
To map this unbounded ratio to a normalized coefficient $\omega_k^{(t)} \in (0, 1)$, we apply the sigmoid function $\omega_k^{(t)} = \sigma(\rho_k^{(t)})$. A larger $\omega_k^{(t)}$ indicates that the stream is learning effective features with high generalization (dominant status), whereas a smaller value indicates optimization stagnation (lagging status). Critically, the validation set is used solely to compute the scalar loss reduction $\Delta\mathcal{L}_{val}^{k,(t)}$ as a read-only signal; no gradients are back-propagated through validation data, and no validation samples participate in parameter updates, which strictly preserves the independence of the validation set. The rationale for amplifying the lagging stream (small $\omega_k$) follows the principle of curriculum learning: a stream that has not yet transferred training knowledge to the validation set remains in an under-explored region of the loss landscape, where increased gradient magnitude promotes feature discovery rather than overfitting.

To balance convergence, we scale the back-propagated gradients based on the relative ranking of these coefficients. Let $k_{min}$ and $k_{max}$ denote the indices of the streams with the minimum and maximum coefficients, respectively. We calculate a deterministic scaling factor $\lambda_k^{(t)}$ for the gradients $\nabla_{\Theta_k}$ as follows:
\begin{equation}
    \lambda_k^{(t)} = 
    \begin{cases} 
    1 + \beta \cdot \omega_k^{(t)}, & \text{if } k = k_{min} \\
    1 - \alpha \cdot \omega_k^{(t)}, & \text{if } k = k_{max}
    \end{cases}
    .
\end{equation}
where $\alpha$ and $\beta$ are hyperparameters that regulate the strength of suppression and enhancement. Because $\omega_k^{(t)} \in (0, 1)$, this piecewise formulation strictly guarantees that the lagging stream receives an enhancement multiplier greater than 1, whereas the dominant stream is penalized with a multiplier less than 1. Finally, the parameters are updated by using the scaled gradients $\nabla_{\Theta_k}' = \lambda_k^{(t)} \cdot \nabla_{\Theta_k}$. This mechanism linearly enhances the updates for the lagging stream while suppressing the updates for the dominant stream. This prevents the model from converging to a biased local optimum driven solely by statistically salient infrastructure metrics, and it forces the framework to capture complex topological dependencies.

\subsection{Optimization Objective}
\begin{figure}[ht]
\centering
\includegraphics[width=0.7\columnwidth]{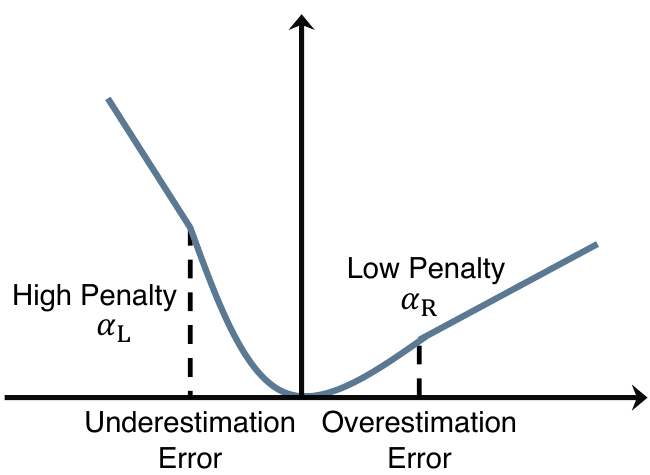}
\caption{Schematic of the Asymmetric Percentage Huber Loss. The function applies a stricter penalty ($\alpha_L$) for underestimation errors to prevent critical SLO violations, while assigning a lower weight ($\alpha_R$) to overestimation.}
\label{fig:loss}
\end{figure}

The final unified embedding $\mathbf{z}_{f}$ is passed through a prediction head based on a multilayer perceptron to produce the latency estimate $\hat{y}$. While prior works, including GRAF~\cite{park2021graf}, use asymmetric squared errors to penalize under-prediction, the purely quadratic formulation in these methods remains highly vulnerable to extreme long-tail outliers. This vulnerability destabilizes the optimization process. To resolve this issue, we use the asymmetric percentage Huber loss, as illustrated in Fig. \ref{fig:loss}. 

We first define the percentage error $e_p$ between the predicted latency and the ground-truth latency as $e_p = (\hat{y} - y) / (y + \epsilon)$. To differentiate prediction costs, we formulate the loss by using a weighted standard Huber function. Let the base Huber term $H(e, \theta)$ be defined as follows:
\begin{equation}
    H(e, \theta) = 
    \begin{cases} 
        e^2, & \text{if } |e| \le \theta \\
        2\theta|e| - \theta^2, & \text{if } |e| > \theta
    \end{cases}
    .
\end{equation}
We then construct the asymmetric objective by applying distinct penalty weights based on the sign of the error:
\begin{equation}
    \mathcal{L}(e_p) = 
    \begin{cases} 
      \alpha_R \cdot H(e_p, \theta_R), & \text{if } e_p \ge 0 \\
      \alpha_L \cdot H(e_p, \theta_L), & \text{if } e_p < 0
    \end{cases}
    .
\end{equation}
As depicted in Fig. \ref{fig:loss}, we strictly enforce $\alpha_L > \alpha_R$. Unlike the unbounded quadratic penalties used in GRAF~\cite{park2021graf}, this configuration mitigates the risk of severe violations of Service Level Objectives through steeper gradients for under-prediction, while it bounds the influence of outliers via linear scaling. This design acknowledges that missing a latency spike is operationally more detrimental than mild over-prediction. Finally, the gradients derived from this objective are scaled by the RAGM mechanism described in Section \ref{sec:ragm} to ensure balanced optimization.

\section{Evaluation}
In this section, we conduct extensive experiments to evaluate the effectiveness of USRFNet. Specifically, we aim to answer the following three research questions:

\textbf{RQ1:} How well does USRFNet perform in P95 tail latency prediction?

\textbf{RQ2:} Does each component contribute to the performance of USRFNet?

\textbf{RQ3:} How do the major hyperparameters of USRFNet influence its performance?

\subsection{Experimental Setup}
\subsubsection{Datasets}
We evaluate our approach on three open-source microservice benchmarks: Online Boutique~\cite{onlineboutique2025}, Sock Shop~\cite{sockshop2025}, and Train Ticket~\cite{zhou2018benchmarking}. The Online Boutique and Sock Shop benchmarks emulate e-commerce platforms with 11 and 13 microservices, respectively, which cover functional components from product catalog browsing to payment processing. To evaluate scalability on a larger system, we use the Train Ticket benchmark, which simulates a comprehensive railway ticketing platform. Comprising 41 microservices, Train Ticket features a deeper dependency graph and more complex service interactions than the e-commerce baselines.

\begin{figure}[t]
    \centering
    \includegraphics[width=0.75\columnwidth]{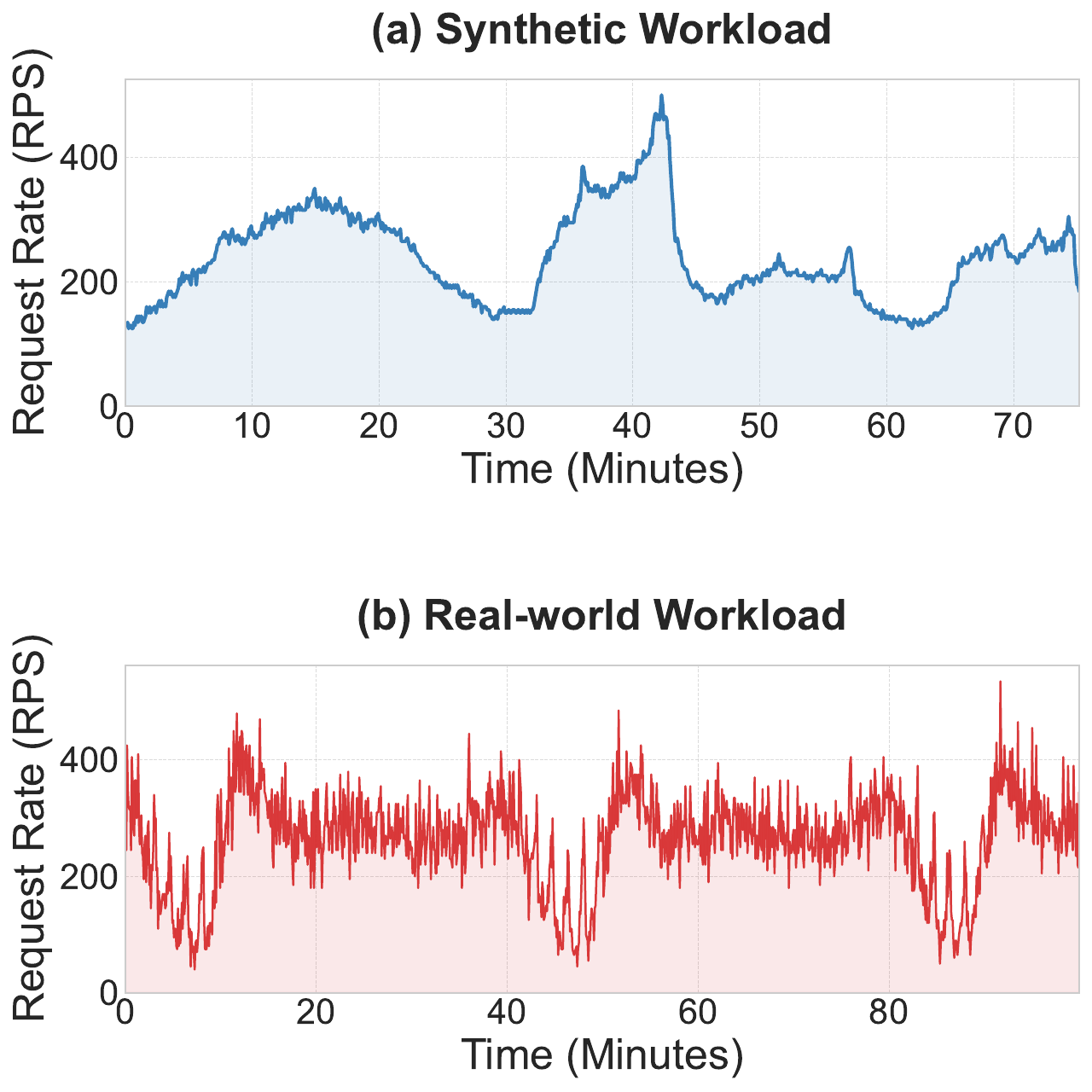}
    \caption{The workload intensity profiles used in the evaluation. (a) The synthetic workload generated by Locust, featuring ramps and spikes to simulate daily cycles and flash crowds. (b) The real-world workload derived from the Alibaba cluster trace, exhibiting complex, non-stationary fluctuations typical of production environments.}
    \label{fig:workload_profile}
\end{figure}

\begin{table*}[t]
\caption{Overall performance evaluation on three benchmarks.}
\centering
\setlength{\tabcolsep}{3pt} 
\begin{tabularx}{\textwidth}{c | *{3}{>{\centering\arraybackslash}X} | *{3}{>{\centering\arraybackslash}X} | *{3}{>{\centering\arraybackslash}X}}
\toprule
\multirow{2}{*}{\textbf{Model}} & \multicolumn{3}{c|}{\textbf{Online Boutique}} & \multicolumn{3}{c|}{\textbf{Sock Shop}} & \multicolumn{3}{c}{\textbf{Train Ticket}} \\ 
\cmidrule{2-10}
 & MAPE(\%) & MAE(s) & RMSE(s) 
 & MAPE(\%) & MAE(s) & RMSE(s) 
 & MAPE(\%) & MAE(s) & RMSE(s) \\
\midrule
GBDT   & 25.16 & 0.077 & 0.117 & 13.36 & 0.037 & 0.077 & 17.18 & 0.107 & 0.139 \\
MLP    & 16.63 & 0.049 & \underline{0.077} & 10.96 & \underline{0.031} & 0.075 & 14.08 & 0.090 & 0.126 \\
FT-Transformer   & 12.38 & 0.041 & 0.086 & 9.95 & 0.040 & 0.124 & \underline{9.22} & \underline{0.064} & 0.105 \\
GRAF   & 14.82 & 0.056 & 0.102 & 9.75  & 0.042 & 0.096 & 12.60 & 0.099 & 0.148 \\
GCN\textsuperscript{+} & 12.19 & 0.041 & 0.081 & 9.18 & 0.033 & 0.076 & 10.78 & 0.072 & 0.106 \\
GIN\textsuperscript{+} & 12.37 & \underline{0.040} & 0.080 & 9.48 & 0.034 & 0.076 & 10.19 & 0.068 & 0.103 \\
GatedGCN\textsuperscript{+} & \underline{11.91} & 0.042 & 0.084 & \underline{9.17} & 0.032 & \underline{0.073} & 9.86 & 0.067 & \underline{0.098} \\
\rowcolor{gray!20}\textbf{USRFNet} & \textbf{8.80} & \textbf{0.030} & \textbf{0.070} & \textbf{7.41} & \textbf{0.026} & \textbf{0.064} & \textbf{7.78} & \textbf{0.054} & \textbf{0.084} \\
\bottomrule
\end{tabularx}
\label{table1}
\end{table*}

To assess the robustness of the model under controlled patterns and unpredictable scenarios, we use a hybrid workload generation strategy. For the e-commerce benchmarks, we generate synthetic traffic using Locust~\cite{locust2025} following a strategy based on prior studies~\cite{meng2023deepscaler}.  This strategy incorporates dynamic intensity profiles that mimic real-world fluctuations (e.g., gradual ramps for daily cycles and sharp spikes for flash crowds), as illustrated in Fig.~\ref{fig:workload_profile}(a). In contrast, the workload for the Train Ticket benchmark is driven by real-world load curves derived from the Alibaba cluster trace~\cite{luo2021characterizing}. As depicted in Fig.~\ref{fig:workload_profile}(b), this trace lacks the regular periodicity of synthetic data and exhibits severe volatility and non-stationary behaviors. It features rapid, irregular oscillations and sustained high-concurrency plateaus. The incorporation of this production-grade variability tests the ability of the model to generalize to the transient load shifts typical of large-scale cloud environments.

These benchmarks are deployed on Kubernetes environments configured for their specific computational demands. The Online Boutique and Sock Shop benchmarks operate on a three-node cluster with 88 vCPUs and 256GB of RAM. The more complex Train Ticket system operates on a six-node cluster equipped with 256 vCPUs and 512GB of RAM. We use Prometheus and Istio to collect fine-grained infrastructure and service-level metrics over a period of five days. To evaluate the robustness of the model against infrastructure volatility, the data collection process includes continuous instance-level dynamics. We introduce this dynamics by randomly executing horizontal pod scaling across the deployment at specific intervals.  This execution ensures that the datasets reflect the structural variability of real-world Kubernetes environments. It tests the ability of the capacity-side encoder to detect sudden shifts in system processing capability and workload redistribution without modifying the static service-level graph backbone. The collected metrics are aggregated by using a sliding window with a 30-second duration and a 5-second step, which ensures high temporal resolution and stability. This process yields extensive datasets with 69,121 samples for Online Boutique, 71,696 samples for Sock Shop, and 85,209 samples for Train Ticket. We chronologically partition each dataset into training (70\%), validation (10\%), and testing (20\%) subsets to simulate realistic deployment scenarios where future data is unseen. Appendix A provides further details regarding the datasets.

\subsubsection{Baselines}

We evaluate USRFNet against several competitive baselines. Traditional models, such as GBDT and MLP, process identical features as flat vectors, which ignores explicit microservice topologies and the physical distinctions between resource metrics and traffic metrics. We include FT-Transformer~\cite{gorishniy2021revisiting} to evaluate advanced deep learning architectures designed for tabular data. To evaluate structural modeling, we compare USRFNet with GNN-based baselines, including the domain-specific GRAF~\cite{park2021graf} and the latest state-of-the-art general-purpose models for graph tasks, such as GCN\textsuperscript{+}, GIN\textsuperscript{+}, and GatedGCN\textsuperscript{+}~\cite{luo2025can}. All graph models receive identical features to ensure a fair comparison. These models represent the single-stream feature fusion paradigm. We exclude per-request models, such as PERT-GNN~\cite{tam2023pert} and FastPert~\cite{xu2025fastpert}, because these models target individual requests rather than the window-level aggregate tail latency required for operational decisions. Furthermore, we omit time-series forecasting models. This study formulates the problem as a snapshot spatial regression rather than a long-term evolutionary prediction. As frameworks such as GRAF~\cite{park2021graf} demonstrate, proactive resource allocation primarily requires estimating the steady-state equilibrium between current system demand and target infrastructure capacity. Because instantaneous tail latency depends on the immediate spatial topology and concurrent resource saturation rather than historical dependencies, time-series architectures remain structurally misaligned with the spatial regression nature of this task. To ensure a fair and isolated evaluation of architectural design choices, all graph-based baselines receive the concatenated union of both traffic and resource features as input, matching the total feature information available to USRFNet. The performance gap therefore reflects differences in architectural inductive bias rather than feature quantity.

\subsubsection{Evaluation Metrics}
We evaluate the prediction performance using three standard statistical metrics: Mean Absolute Percentage Error (MAPE), Mean Absolute Error (MAE), and Root Mean Square Error (RMSE). Let $y_i$ denote the actual ground-truth P95 latency and $\hat{y}_i$ represent the predicted value for the $i$-th sample, where $n$ is the total number of samples in the test set. These metrics are defined as: 
MAPE $= \frac{100\%}{n} \sum_{i=1}^{n} | \frac{y_i - \hat{y}_i}{y_i} |$, 
MAE $= \frac{1}{n} \sum_{i=1}^{n} |y_i - \hat{y}_i|$, and 
RMSE $= \sqrt{\frac{1}{n} \sum_{i=1}^{n} (y_i - \hat{y}_i)^2}$. 
Lower values across all three metrics indicate higher prediction accuracy.

\subsubsection{Implementation Details}
All experiments are conducted on a server equipped with a 12-vCPU Intel Xeon Platinum 8352V processor, 90GB of RAM, and a single NVIDIA RTX 4090 GPU. We implement the proposed model using PyTorch 2.0.0 within an Ubuntu 20.04 environment with CUDA 11.8 support. The hyperparameters for all baselines are tuned to their optimal settings to ensure a fair comparison.

\subsection{RQ1: Overall Performance Evaluation}
Table~\ref{table1} summarizes the predictive performance of USRFNet and all competitive baselines. The results indicate that USRFNet consistently achieves the highest accuracy across all evaluation metrics, which significantly outperforms traditional machine learning models, advanced tabular deep learning architectures, and state-of-the-art graph neural networks.

The performance gains of USRFNet become evident in the MAPE reductions. Specifically, compared to the best-performing baselines, USRFNet achieves relative MAPE reductions of $26.11\%$, $19.19\%$, and $15.62\%$ on the Online Boutique, Sock Shop, and Train Ticket benchmarks, respectively. Although tabular models, such as GBDT, MLP, and the deep learning-based FT-Transformer, use identical features, the inability of these models to account for service topologies and semantic distinctions between resource data and traffic data leads to suboptimal predictions. This limitation confirms that flat feature representations are insufficient to capture the structural complexity of microservice systems. Furthermore, USRFNet consistently outperforms GNN-based baselines such as GRAF and GatedGCN\textsuperscript{+}. Although these graph models incorporate structural information, the associated single-stream fusion paradigm applies a uniform inductive bias that homogenizes the distinct signals of resource utilization and traffic intensity. These results validate that decoupling heterogeneous data streams through a dual-stream architecture prevents interference between distinct signals, which remains essential for accurately modeling tail latency in dynamic cloud environments. An extended evaluation against additional baselines is detailed in Appendix~B1.

\begin{table}[h]
\centering
\caption{Computational Efficiency and Memory Usage of USRFNet.}
\label{table:efficiency}
\setlength{\tabcolsep}{10pt}
\begin{tabular}{l r r}
\toprule
\textbf{Benchmark} & \textbf{Memory Usage} & \textbf{Inference Latency} \\
\midrule
Online Boutique     & 555 KB               & $2.03 \pm 1.13$ ms             \\
Sock Shop           & 552 KB               & $1.93 \pm 1.48$ ms             \\
Train Ticket        & 761 KB               & $2.19 \pm 1.65$ ms             \\
\bottomrule
\end{tabular}
\end{table}

To evaluate the practical deployability of USRFNet, we analyze the computational efficiency and memory requirements of the model. As Table~\ref{table:efficiency} shows, USRFNet maintains a minimal memory footprint across different system scales and achieves minimal inference latency, measured by the average CPU time per observation window. This processing speed results from using instantaneous snapshot regression instead of heavy auto-regressive sequence modeling. Because the metric collection interval is five seconds, an inference time in the millisecond range is effectively negligible. This efficiency allows the prediction process to operate within short control loops. It provides real-time reliability assessments for downstream tasks, such as proactive autoscaling, without significant delay, which meets the real-time requirements of large-scale microservice management.

\begin{table*}[t]
\centering
\caption{Evaluation results of the ablation study across three benchmarks. We evaluate nine variants (C1--C9) to assess the impact of the dual-modality input, sub-network architectures, and core innovation modules of USRFNet.} 
\label{table:ablation_results}
\setlength{\tabcolsep}{3pt}
\begin{tabularx}{\textwidth}{c | *{3}{>{\centering\arraybackslash}X} | *{3}{>{\centering\arraybackslash}X} | *{3}{>{\centering\arraybackslash}X}}
\toprule
\multirow{2}{*}{\textbf{Variant}} & \multicolumn{3}{c|}{\textbf{Online Boutique}} & \multicolumn{3}{c|}{\textbf{Sock Shop}} & \multicolumn{3}{c}{\textbf{Train Ticket}} \\
\cmidrule{2-10}
 & MAPE(\%) & MAE(s) & RMSE(s) & MAPE(\%) & MAE(s) & RMSE(s) & MAPE(\%) & MAE(s) & RMSE(s) \\
\midrule
C1 & 32.12 & 0.143 & 0.254 & 24.51 & 0.106 & 0.246 & 15.38 & 0.109 & 0.160 \\
C2 & 10.75 & 0.037 & 0.076 & 9.06 & 0.032 & 0.076 & 13.97 & 0.101 & 0.149 \\
\midrule
C3 & 16.84 & 0.060 & 0.110 & 10.08 & 0.035 & 0.084 & 9.65 & 0.071 & 0.114 \\
C4 & 13.05 & 0.046 & 0.097 & 9.40  & 0.036 & 0.079 & 11.23 & 0.083 & 0.129 \\
C5 & 10.26 & 0.035 & \underline{0.071} & 8.72  & 0.030 & 0.073 & 12.66 & 0.092 & 0.139 \\
C6 & 9.75 & 0.034 & 0.080 & 8.61  & 0.032 & 0.082 & 10.47 & 0.072 & 0.109 \\
C7 & \underline{9.22} & 0.032 & 0.079 & 8.17  & \underline{0.028} & 0.073 & 10.30 & 0.070 & 0.106 \\
\midrule
C8 & 11.31 & 0.039 & 0.080 & 8.65  & 0.030 & 0.072 & 12.39 & 0.091 & 0.138 \\
C9 & 9.31 & \underline{0.031} & 0.075 & \underline{8.05} & 0.029 & \underline{0.070} & \underline{8.32} & \underline{0.058} & \underline{0.092} \\
\midrule
\rowcolor{gray!20}\textbf{USRFNet} & \textbf{8.80} & \textbf{0.030} & \textbf{0.070} & \textbf{7.41} & \textbf{0.026} & \textbf{0.064} & \textbf{7.78} & \textbf{0.054} & \textbf{0.084} \\
\bottomrule
\end{tabularx}
\end{table*}

\subsection{RQ2: Ablation Study}

To evaluate the contribution of each component, we evaluate nine variants (C1--C9) in Table~\ref{table:ablation_results}. 

We first assess the dual-stream architecture against traffic stream-only (C1) and resource stream-only (C2) baselines. In simpler systems such as Online Boutique, C2 significantly outperforms C1, which confirms resource availability as the primary bottleneck. However, in the highly complex Train Ticket benchmark, this error gap narrows considerably. This modality shift indicates that as topologies deepen, traffic stream propagation becomes as critical as resource stream contention. USRFNet consistently outperforms both variants, confirming the need to integrate heterogeneous data streams.

Next, we validate the proposed sub-network architectures and the fusion paradigm. On the resource stream, replacing the gMLP encoder with GATv2 (C3) or FT-Transformer (C4) severely degrades the prediction performance. GATv2~\cite{brody2022how} restricts feature aggregation to explicit topological edges, which misses implicit global contention (e.g., shared hosts), whereas FT-Transformer~\cite{gorishniy2021revisiting} lacks the multiplicative spatial gating required to effectively isolate capacity bottlenecks. On the traffic stream, GraphSAGE (C6) uniformly aggregates neighboring nodes and ignores edge attributes, whereas GATv2 (C5) treats edge features merely as additive biases. In contrast, the proposed transformer-based graph encoder intrinsically maps edge features into the key space. This structural design enables a direct multiplicative interaction between the traffic volume and the queried service state, dynamically amplifying the attention weights for high-throughput dependencies. Furthermore, standard cross-attention (C7) provides only linear alignment, which fails to capture the inherently multiplicative demand-capacity dynamics modeled by the low-rank tensor fusion of the HIDAC module.

Finally, omitting Cross-Diffusion-Attention (C8) forces the direct fusion of different semantic spaces. This omission prevents the two streams from mutually querying to identify demand-capacity misalignments, which causes the prediction errors to increase. Disabling the RAGM strategy (C9) consistently reduces prediction accuracy, particularly on complex datasets, demonstrating the necessity of this strategy for mitigating gradient interference during joint representation learning. Overall, the consistent performance degradation across variants C1 through C9 confirms that each component improves the overall accuracy of USRFNet. A comprehensive ablation analysis involving additional variants is provided in Appendix~B2.

\subsection{RQ3: Hyperparameter Sensitivity}
\begin{figure}[!ht]
\centering
\includegraphics[width=0.95\columnwidth]{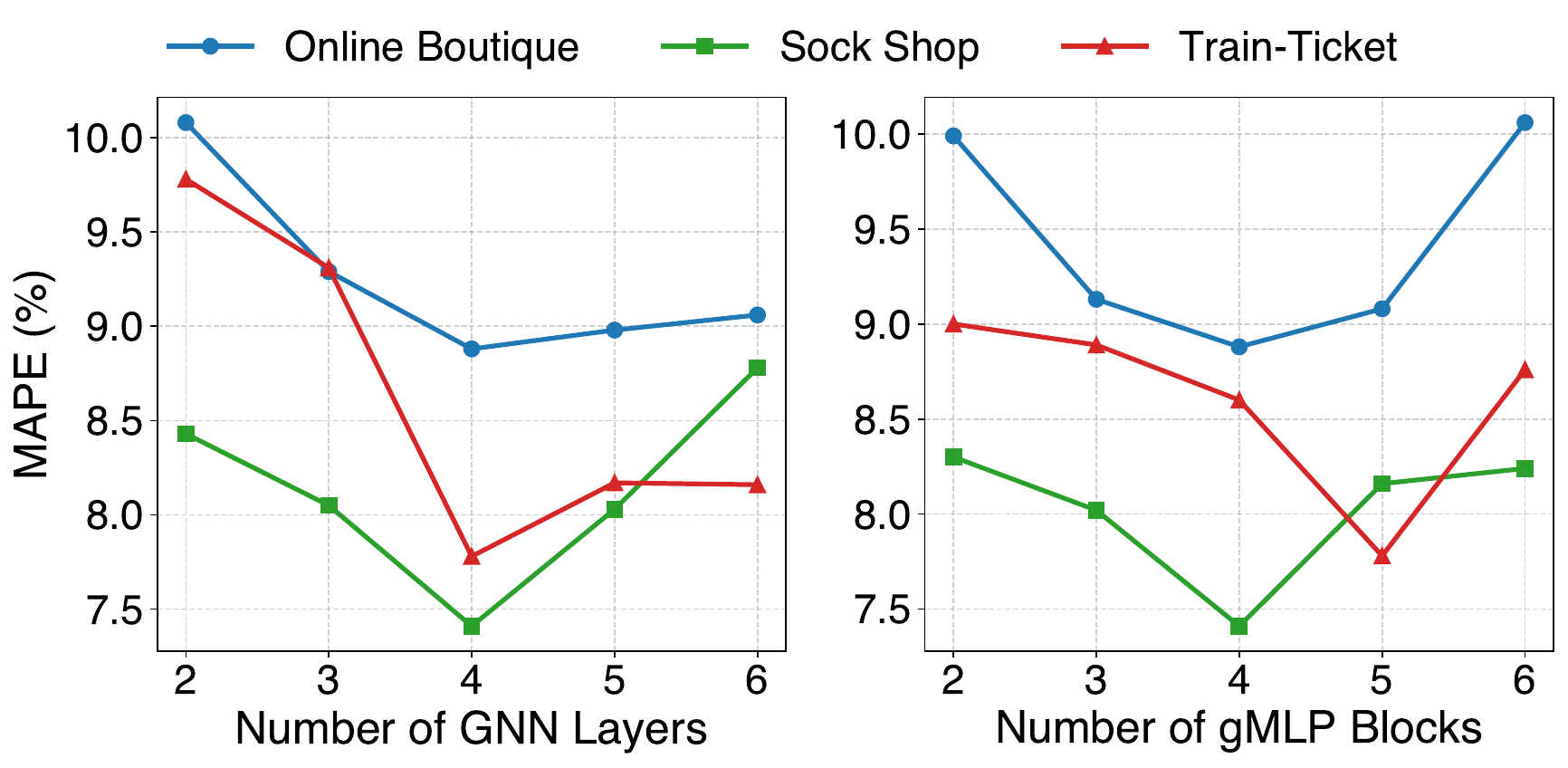}
\caption{Performance sensitivity of USRFNet to the number of GNN layers (left) and gMLP blocks (right) across the three benchmarks.}
\label{fig:architecture_sensitivity}
\end{figure}

We evaluate the sensitivity of USRFNet to the architectural depth of the encoders to identify the optimal configuration for feature extraction. As illustrated in Fig. \ref{fig:architecture_sensitivity}, we vary the number of GNN layers and gMLP blocks from 2 to 6 across the three benchmarks. For the traffic representation module, a depth of four GNN layers achieves the best performance across all datasets. This depth provides a sufficient receptive field to capture multi-hop service invocation chains while avoiding the oversmoothing of node representations. For the resource representation module, a configuration of four gMLP blocks yields the lowest prediction error for the Online Boutique and Sock Shop benchmarks. However, the more complex Train Ticket benchmark requires five blocks to effectively model the complex resource contention patterns. These results indicate that shallow configurations lack the capacity to capture non-linear system dynamics, whereas excessive depth causes overfitting and degrades the prediction accuracy.

\section{Discussion}
\subsection{Case Study}
\label{sec:case_study}

\begin{figure}[!ht]
\centering
\includegraphics[width=0.85\columnwidth]{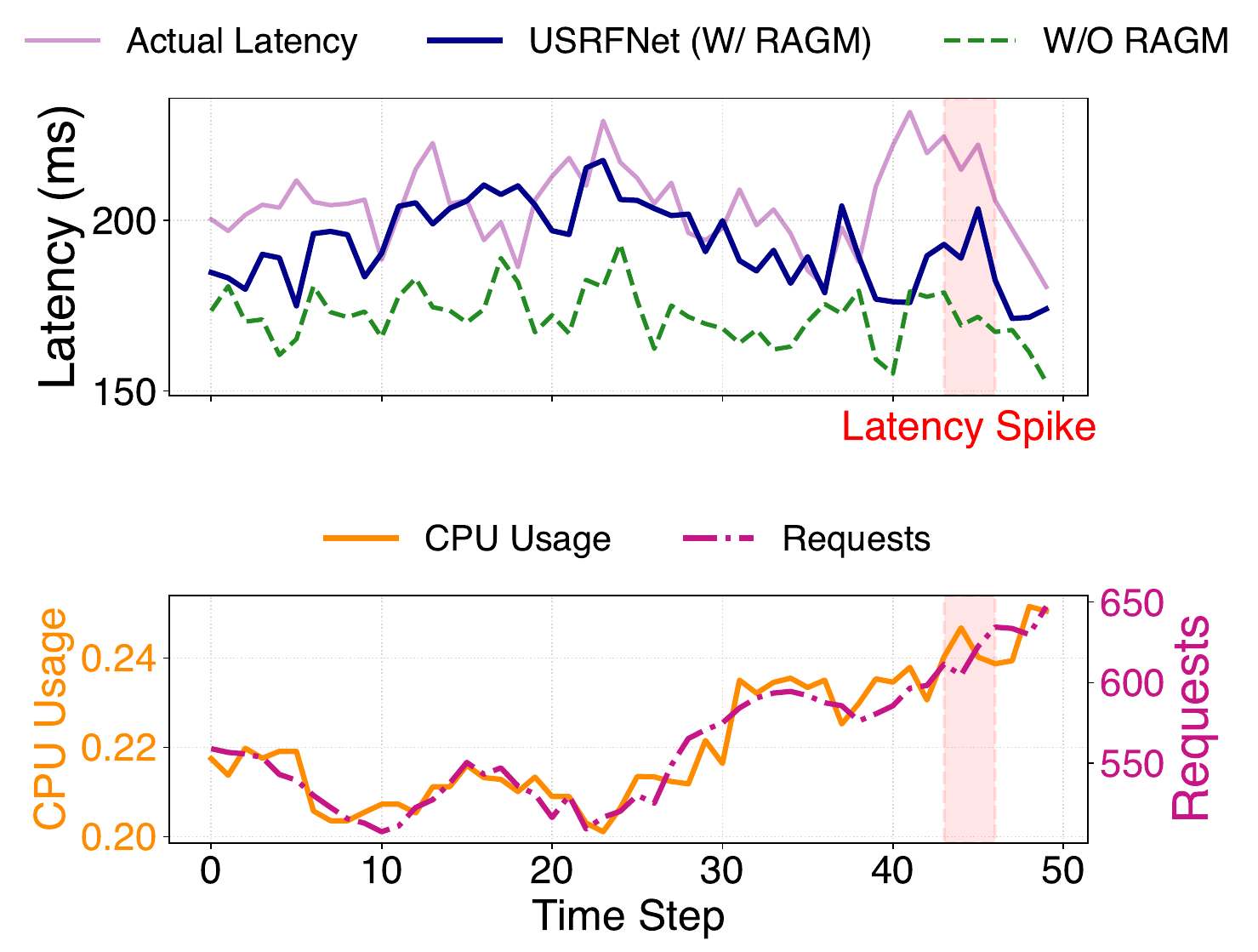}
\caption{Case study on the efficacy of gradient modulation. The bottom panel illustrates the trends of CPU utilization and request volume. The top panel compares the ground-truth latency with predictions from USRFNet and a variant without RAGM. The model without RAGM underestimates the spike because of an over-reliance on resource metrics, whereas the proposed framework successfully captures the traffic-induced latency surge.}
\label{case}
\end{figure}

A high-volatility period in the Online Boutique benchmark demonstrates the necessity of the Reliability-Aware Gradient Modulation mechanism. Fig. \ref{case} compares the prediction trajectories alongside the corresponding system metrics. As illustrated in the bottom panel, simultaneous increases in workload requests and CPU utilization at time step 25 cause a latency spike that exceeds 220 ms.

The performance of the model variants diverges significantly during this critical latency surge, which is indicated by the red shaded region. The variant without modulation fails to capture this spike, and the corresponding prediction remains in a lower range. This failure demonstrates that the optimizer overfits to the fast-converging resource stream, which causes the model to miss the non-linear queuing effects induced by the traffic surge. In contrast, the complete architecture of USRFNet accurately tracks the latency peak. By dynamically scaling the gradients of the lagging traffic stream, the modulation mechanism prevents optimization bias toward the dominant resource features. This modulation ensures effective signal integration for stable predictions under extreme system stress.

\subsection{Threats to Validity}
\label{sec:threats}

A key threat to external validity involves the generalization of USRFNet across diverse infrastructure scales and workloads. To address this threat, the evaluation of the model spans different cluster configurations, ranging from a standard three-node setup (88 vCPUs, 256GB RAM) to a six-node environment (256 vCPUs, 512GB RAM). Furthermore, the evaluation incorporates distinct traffic patterns, including synthetic loads with controlled fluctuations and an Alibaba cluster trace featuring industrial-grade volatility. This comprehensive evaluation confirms that the performance improvements hold across hardware and workload variations. Regarding internal validity, the performance of the model depends on the configuration of the RAGM hyperparameters $\alpha$ and $\beta$. Since the optimal parameter configurations may vary across different infrastructure setups, please refer to Appendix~B4 for a detailed analysis of the model robustness. Finally, construct validity is maintained through the use of standard evaluation metrics, including MAPE, MAE, and RMSE.

\subsection{Potential for System-Level Integration}

\begin{figure}[!ht]
\centering
\includegraphics[width=0.85\columnwidth]{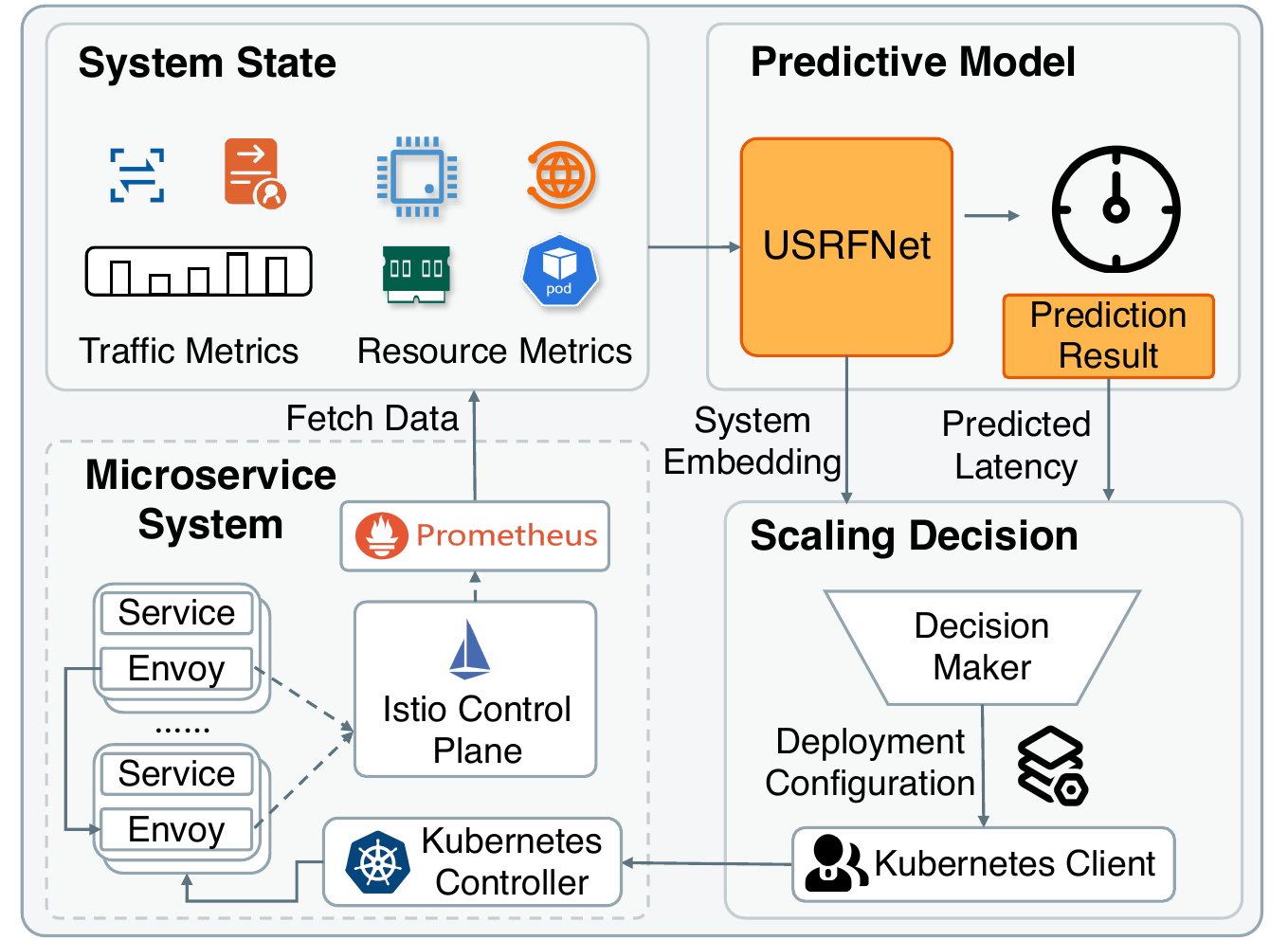} 
\caption{Conceptual integration of USRFNet into the autonomic control loop. The framework provides the predictive foundation for proactive resource orchestration and generates system embeddings to support contextual decision-making within the management pipeline.}
\label{Application}
\end{figure}

USRFNet functions as a predictive engine within the autonomic management loop of microservice systems, as illustrated in Fig.~\ref{Application}.  Within the MAPE-K framework, USRFNet occupies the analysis phase to transform raw telemetry data into proactive insights. The model provides two primary capabilities for software performance engineering. The prediction of window-level P95 latency provides a stable indicator to anticipate violations of Service Level Objectives, which allows downstream components, such as autoscalers, to adjust resources before the performance degrades. Furthermore, the system embeddings compactly represent the interaction between system demand and infrastructure capacity. These representations support complex tasks, such as root cause analysis and policy optimization, more effectively than scalar metrics.

In practice, USRFNet integrates as a middle-tier inference service. The framework processes telemetry data collected from monitoring frameworks such as Prometheus~\cite{prometheus2025} and service meshes such as Istio~\cite{istio2025} for real-time inference. Although downstream control policies are beyond the scope of this study, the accuracy of USRFNet provides a reliable foundation for these autonomic functions. By providing a forecast aware of Service Level Objectives, USRFNet supports the transition from reactive maintenance to proactive performance management in cloud environments.

\section{Related Work}
\label{sec:related_works}

Microservice performance estimation is essential for tasks such as distributed tracing~\cite{he2023steam, huang2025mint}, root cause analysis~\cite{somashekar2024gamma, lin2024root, pham2024root}, and proactive resource scaling~\cite{luo2022erms, luo2022power, meng2023deepscaler, chen2024derm, chen2025grad}. Accurate tail latency prediction represents the fundamental prerequisite for these applications. Existing predictive models follow three primary paradigms.

\paragraph{Analytical and Machine Learning Approaches}
Early models formalize microservice behavior using queuing networks or Markov models~\cite{gias2019atom, bhasi2021kraken, luo2022erms}, which rely on rigid assumptions regarding service time distributions. To relax these constraints, data-driven methods use classical machine learning~\cite{gan2019seer, yang2019miras, zhang2021sinan} or causal inference~\cite{gan2021sage, chow2022deeprest}. However, by aggregating system telemetry into flat vectors, these approaches lack topological awareness. They fail to capture cascading delays and multi-hop queuing effects, which results in suboptimal predictions.

\paragraph{Per-Request Tracing Models}
To incorporate structural dependencies, trace-level models such as PERT-GNN~\cite{tam2023pert} and FastPERT~\cite{xu2025fastpert} use Graph Neural Networks to predict the latency of individual requests. Although suitable for offline diagnostics, processing massive volumes of per-request traces incurs substantial computational overhead. Therefore, these fine-grained models are impractical for the real-time, window-level aggregate tail latency prediction required for proactive autoscaling.

\paragraph{Single-Stream Graph Models}
For efficient system-level prediction, methods such as GRAF~\cite{park2021graf} estimate aggregated metrics over fixed time intervals using macro-topology graphs. However, they adopt a single-stream feature fusion paradigm that mixes highly heterogeneous traffic signals and resource signals through uniform message passing. This approach introduces structural biases by constraining implicit global resource contention with explicit invocation edges. Furthermore, the reliance on simple concatenation fails to capture the non-linear, multiplicative dynamics between workload demand and infrastructure capacity.

In contrast, USRFNet introduces a dual-stream architecture that strictly decouples these heterogeneous modalities. Instead of prematurely combining signals, the proposed framework explicitly separates the modeling of topological propagation and global resource contention before integrating them through a hierarchical fusion mechanism. This design effectively overcomes the structural biases and semantic homogenization present in existing methods to enable a more expressive modeling of the system.

\section{Conclusion} 
In this paper, we propose USRFNet, a novel dual-stream architecture designed for accurate window-level tail latency prediction in microservices. To address the limitations of existing single-stream approaches, USRFNet explicitly decouples the modeling of heterogeneous telemetry data. Specifically, it employs a tailored GNN to capture traffic stream topological propagation and a gMLP to distill resource stream global contention. Furthermore, these distinct representations are effectively integrated through the Hierarchical Integration of Demand and Capacity (HIDAC) module, and the joint training is optimized using the Reliability-Aware Gradient Modulation (RAGM) strategy to mitigate inter-modal interference. Extensive experiments on three real-world benchmarks demonstrate that USRFNet significantly outperforms state-of-the-art models, validating the necessity of specialized multi-modality modeling. In the future, we plan to incorporate lightweight sequence modeling for temporal forecasting and integrate path-aware embeddings into proactive resource orchestration frameworks to ensure highly reliable and cost-effective cloud-native operations.

\section*{Acknowledgments}
We acknowledge the use of ChatGPT, a large language model developed by OpenAI, to polish the English grammar and improve the readability of the manuscript (excluding the references section). The authors thoroughly reviewed and edited the AI-assisted text as needed and take full responsibility for the final content of this paper.



\bibliographystyle{IEEEtran} 
\bibliography{IEEEabrv,ref} 

\begin{thebibliography}{10}
\providecommand{\url}[1]{#1}
\csname url@samestyle\endcsname
\providecommand{\newblock}{\relax}
\providecommand{\bibinfo}[2]{#2}
\providecommand{\BIBentrySTDinterwordspacing}{\spaceskip=0pt\relax}
\providecommand{\BIBentryALTinterwordstretchfactor}{4}
\providecommand{\BIBentryALTinterwordspacing}{\spaceskip=\fontdimen2\font plus
\BIBentryALTinterwordstretchfactor\fontdimen3\font minus \fontdimen4\font\relax}
\providecommand{\BIBforeignlanguage}[2]{{%
\expandafter\ifx\csname l@#1\endcsname\relax
\typeout{** WARNING: IEEEtran.bst: No hyphenation pattern has been}%
\typeout{** loaded for the language `#1'. Using the pattern for}%
\typeout{** the default language instead.}%
\else
\language=\csname l@#1\endcsname
\fi
#2}}
\providecommand{\BIBdecl}{\relax}
\BIBdecl

\bibitem{dragoni2017microservices}
N.~Dragoni, S.~Giallorenzo, A.~L. Lafuente, M.~Mazzara, F.~Montesi, R.~Mustafin, and L.~Safina, ``Microservices: yesterday, today, and tomorrow,'' \emph{Present and ulterior software engineering}, pp. 195--216, 2017.

\bibitem{sriraman2019softsku}
A.~Sriraman, A.~Dhanotia, and T.~F. Wenisch, ``Softsku: Optimizing server architectures for microservice diversity@ scale,'' in \emph{Proceedings of the 46th International Symposium on Computer Architecture}, 2019, pp. 513--526.

\bibitem{liu2021microhecl}
D.~Liu, C.~He, X.~Peng, F.~Lin, C.~Zhang, S.~Gong, Z.~Li, J.~Ou, and Z.~Wu, ``Microhecl: High-efficient root cause localization in large-scale microservice systems,'' in \emph{2021 IEEE/ACM 43rd International Conference on Software Engineering: Software Engineering in Practice (ICSE-SEIP)}, 2021, pp. 338--347.

\bibitem{onlineboutique2025}
G.~C. Platform, ``Online boutique,'' https://github.com/GoogleCloudPlatform/microservices-demo, 2025.

\bibitem{tam2023pert}
D.~S.~H. Tam, Y.~Liu, H.~Xu, S.~Xie, and W.~C. Lau, ``Pert-gnn: Latency prediction for microservice-based cloud-native applications via graph neural networks,'' in \emph{Proceedings of the 29th ACM SIGKDD Conference on Knowledge Discovery and Data Mining}, 2023, pp. 2155--2165.

\bibitem{xu2025fastpert}
D.~S.~H. Tam, H.~Xu, Y.~Liu, S.~Xie, and W.~C. Lau, ``Fastpert: Towards fast microservice application latency prediction via structural inductive bias over pert networks,'' in \emph{Proceedings of the AAAI Conference on Artificial Intelligence}, 2025, pp. 20\,787--20\,795.

\bibitem{park2021graf}
J.~Park, B.~Choi, C.~Lee, and D.~Han, ``Graph neural network-based slo-aware proactive resource autoscaling framework for microservices,'' \emph{IEEE/ACM Transactions on Networking}, vol.~32, no.~4, pp. 3331--3346, 2024.

\bibitem{somashekar2024gamma}
G.~Somashekar, A.~Dutt, M.~Adak, T.~Lorido~Botran, and A.~Gandhi, ``Gamma: Graph neural network-based multi-bottleneck localization for microservices applications,'' in \emph{Proceedings of the ACM Web Conference 2024}, 2024, pp. 3085--3095.

\bibitem{sun2024art}
Y.~Sun, B.~Shi, M.~Mao, M.~Ma, S.~Xia, S.~Zhang, and D.~Pei, ``Art: A unified unsupervised framework for incident management in microservice systems,'' in \emph{Proceedings of the 39th IEEE/ACM International Conference on Automated Software Engineering}, 2024, pp. 1183--1194.

\bibitem{luo2022erms}
S.~Luo, H.~Xu, K.~Ye, G.~Xu, L.~Zhang, J.~He, G.~Yang, and C.~Xu, ``Erms: Efficient resource management for shared microservices with sla guarantees,'' in \emph{Proceedings of the 28th ACM International Conference on Architectural Support for Programming Languages and Operating Systems, Volume 1}, 2022, pp. 62--77.

\bibitem{peng2022balanced}
X.~Peng, Y.~Wei, A.~Deng, D.~Wang, and D.~Hu, ``Balanced multimodal learning via on-the-fly gradient modulation,'' in \emph{Proceedings of the IEEE/CVF conference on computer vision and pattern recognition}, 2022, pp. 8238--8247.

\bibitem{scarselli2008graph}
F.~Scarselli, M.~Gori, A.~C. Tsoi, M.~Hagenbuchner, and G.~Monfardini, ``The graph neural network model,'' \emph{IEEE transactions on neural networks}, vol.~20, no.~1, pp. 61--80, 2008.

\bibitem{shi2021masked}
Y.~Shi, Z.~Huang, S.~Feng, H.~Zhong, W.~Wang, and Y.~Sun, ``Masked label prediction: Unified message passing model for semi-supervised classification,'' in \emph{Proceedings of the Thirtieth International Joint Conference on Artificial Intelligence}, 2021, pp. 1548--1554.

\bibitem{liu2021pay}
H.~Liu, Z.~Dai, D.~So, and Q.~V. Le, ``Pay attention to mlps,'' in \emph{Advances in neural information processing systems}, 2021, pp. 9204--9215.

\bibitem{sockshop2025}
WeaveWorks, ``Sock shop : A microservice demo application,'' https://github.com/ocp-power-demos/sock-shop-demo, 2025.

\bibitem{zhou2018benchmarking}
X.~Zhou, X.~Peng, T.~Xie, J.~Sun, C.~Xu, C.~Ji, and W.~Zhao, ``Benchmarking microservice systems for software engineering research,'' in \emph{Proceedings of the 40th International Conference on Software Engineering: Companion Proceedings}, 2018, pp. 323--324.

\bibitem{tao2024giving}
L.~Tao, S.~Zhang, Z.~Jia, J.~Sun, M.~Ma, Z.~Li, Y.~Sun, C.~Yang, Y.~Zhang, and D.~Pei, ``Giving every modality a voice in microservice failure diagnosis via multimodal adaptive optimization,'' in \emph{Proceedings of the 39th IEEE/ACM International Conference on Automated Software Engineering}, 2024, pp. 1107--1119.

\bibitem{huang2025adaptive}
C.~Huang, Y.~Wei, Z.~Yang, and D.~Hu, ``Adaptive unimodal regulation for balanced multimodal information acquisition,'' in \emph{Proceedings of the Computer Vision and Pattern Recognition Conference}, 2025, pp. 25\,854--25\,863.

\bibitem{kipf2017semisupervised}
\BIBentryALTinterwordspacing
T.~N. Kipf and M.~Welling, ``Semi-supervised classification with graph convolutional networks,'' in \emph{International Conference on Learning Representations}, 2017. [Online]. Available: \url{https://openreview.net/forum?id=SJU4ayYgl}
\BIBentrySTDinterwordspacing

\bibitem{veličković2018graph}
\BIBentryALTinterwordspacing
P.~Veličković, G.~Cucurull, A.~Casanova, A.~Romero, P.~Liò, and Y.~Bengio, ``Graph attention networks,'' in \emph{International Conference on Learning Representations}, 2018. [Online]. Available: \url{https://openreview.net/forum?id=rJXMpikCZ}
\BIBentrySTDinterwordspacing

\bibitem{gorishniy2021revisiting}
Y.~Gorishniy, I.~Rubachev, V.~Khrulkov, and A.~Babenko, ``Revisiting deep learning models for tabular data,'' in \emph{Advances in neural information processing systems}, vol.~34, 2021, pp. 18\,932--18\,943.

\bibitem{vaswani2017attention}
A.~Vaswani, N.~Shazeer, N.~Parmar, J.~Uszkoreit, L.~Jones, A.~N. Gomez, {\L}.~Kaiser, and I.~Polosukhin, ``Attention is all you need,'' in \emph{Advances in neural information processing systems}, vol.~30, 2017.

\bibitem{wang2024mutualformer}
X.~Wang, X.~Wang, B.~Jiang, J.~Tang, and B.~Luo, ``Mutualformer: Multi-modal representation learning via cross-diffusion attention,'' \emph{International Journal of Computer Vision}, vol. 132, no.~9, pp. 3867--3888, 2024.

\bibitem{liu2018efficient}
Z.~Liu and Y.~Shen, ``Efficient low-rank multimodal fusion with modality-specific factors,'' in \emph{Proceedings of the 56th Annual Meeting of the Association for Computational Linguistics (Long Papers)}, 2018, pp. 2247--2256.

\bibitem{locust2025}
H.~Jonatan, B.~Carl, H.~Joakim, and H.~Heyman, ``Locust,'' https://locust.io/, 2025.

\bibitem{meng2023deepscaler}
C.~Meng, S.~Song, H.~Tong, M.~Pan, and Y.~Yu, ``Deepscaler: Holistic autoscaling for microservices based on spatiotemporal gnn with adaptive graph learning,'' in \emph{2023 38th IEEE/ACM International Conference on Automated Software Engineering (ASE)}, 2023, pp. 53--65.

\bibitem{luo2021characterizing}
S.~Luo, H.~Xu, C.~Lu, K.~Ye, G.~Xu, L.~Zhang, Y.~Ding, J.~He, and C.~Xu, ``Characterizing microservice dependency and performance: Alibaba trace analysis,'' in \emph{Proceedings of the ACM symposium on cloud computing}, 2021, pp. 412--426.

\bibitem{luo2025can}
Y.~Luo, L.~Shi, and X.-M. Wu, ``Can classic {GNN}s be strong baselines for graph-level tasks? simple architectures meet excellence,'' in \emph{Forty-second International Conference on Machine Learning}, 2025, pp. 41\,290--41\,310.

\bibitem{brody2022how}
\BIBentryALTinterwordspacing
B.~Shaked, A.~Uri, and Y.~Eran, ``How attentive are graph attention networks?'' in \emph{International Conference on Learning Representations}, 2022. [Online]. Available: \url{https://openreview.net/forum?id=F72ximsx7C1}
\BIBentrySTDinterwordspacing

\bibitem{prometheus2025}
SoundCloud, ``Prometheus,'' https://prometheus.io/, 2025.

\bibitem{istio2025}
T.~Varun and R.~Louis, ``Istio,'' https://istio.io/, 2026.

\bibitem{he2023steam}
S.~He, B.~Feng, L.~Li, X.~Zhang, Y.~Kang, Q.~Lin, S.~Rajmohan, and D.~Zhang, ``Steam: Observability-preserving trace sampling,'' in \emph{Proceedings of the 31st ACM Joint European Software Engineering Conference and Symposium on the Foundations of Software Engineering}, 2023, pp. 1750--1761.

\bibitem{huang2025mint}
H.~Huang, C.~Chen, K.~Chen, P.~Chen, G.~Yu, Z.~He, Y.~Wang, H.~Zhang, and Q.~Zhou, ``Mint: Cost-efficient tracing with all requests collection via commonality and variability analysis,'' in \emph{Proceedings of the 30th ACM International Conference on Architectural Support for Programming Languages and Operating Systems, Volume 1}, 2025, pp. 683--697.

\bibitem{lin2024root}
C.-M. Lin, C.~Chang, W.-Y. Wang, K.-D. Wang, and W.-C. Peng, ``Root cause analysis in microservice using neural granger causal discovery,'' in \emph{Proceedings of the AAAI Conference on Artificial Intelligence}, 2024, pp. 206--213.

\bibitem{pham2024root}
L.~Pham, H.~Ha, and H.~Zhang, ``Root cause analysis for microservice system based on causal inference: How far are we?'' in \emph{Proceedings of the 39th IEEE/ACM International Conference on Automated Software Engineering}, 2024, pp. 706--715.

\bibitem{luo2022power}
S.~Luo, H.~Xu, K.~Ye, G.~Xu, L.~Zhang, G.~Yang, and C.~Xu, ``The power of prediction: microservice auto scaling via workload learning,'' in \emph{Proceedings of the 13th Symposium on Cloud Computing}, 2022, pp. 355--369.

\bibitem{chen2024derm}
L.~Chen, S.~Luo, C.~Lin, Z.~Mo, H.~Xu, K.~Ye, and C.~Xu, ``Derm: Sla-aware resource management for highly dynamic microservices,'' in \emph{2024 ACM/IEEE 51st Annual International Symposium on Computer Architecture (ISCA)}, 2024, pp. 424--436.

\bibitem{chen2025grad}
L.~Chen, C.~Lin, S.~Luo, H.~Xu, and C.~Xu, ``Grad: Intelligent microservice scaling by harnessing resource fungibility,'' in \emph{2025 IEEE International Symposium on High Performance Computer Architecture}, 2025, pp. 474--486.

\bibitem{gias2019atom}
A.~U. Gias, G.~Casale, and M.~Woodside, ``Atom: Model-driven autoscaling for microservices,'' in \emph{2019 IEEE 39th International Conference on Distributed Computing Systems (ICDCS)}, 2019, pp. 1994--2004.

\bibitem{bhasi2021kraken}
V.~M. Bhasi, J.~R. Gunasekaran, P.~Thinakaran, C.~S. Mishra, M.~T. Kandemir, and C.~Das, ``Kraken: Adaptive container provisioning for deploying dynamic dags in serverless platforms,'' in \emph{Proceedings of the ACM Symposium on Cloud Computing}, 2021, pp. 153--167.

\bibitem{gan2019seer}
Y.~Gan, Y.~Zhang, K.~Hu, D.~Cheng, Y.~He, M.~Pancholi, and C.~Delimitrou, ``Seer: Leveraging big data to navigate the complexity of performance debugging in cloud microservices,'' in \emph{Proceedings of the twenty-fourth international conference on architectural support for programming languages and operating systems}, 2019, pp. 19--33.

\bibitem{yang2019miras}
Z.~Yang, P.~Nguyen, H.~Jin, and K.~Nahrstedt, ``Miras: Model-based reinforcement learning for microservice resource allocation over scientific workflows,'' in \emph{2019 IEEE 39th international conference on distributed computing systems (ICDCS)}, 2019, pp. 122--132.

\bibitem{zhang2021sinan}
Y.~Zhang, W.~Hua, Z.~Zhou, G.~E. Suh, and C.~Delimitrou, ``Sinan: Ml-based and qos-aware resource management for cloud microservices,'' in \emph{Proceedings of the 26th ACM international conference on architectural support for programming languages and operating systems}, 2021, pp. 167--181.

\bibitem{gan2021sage}
Y.~Gan, M.~Liang, S.~Dev, D.~Lo, and C.~Delimitrou, ``Sage: practical and scalable ml-driven performance debugging in microservices,'' in \emph{Proceedings of the 26th ACM International Conference on Architectural Support for Programming Languages and Operating Systems}, 2021, pp. 135--151.

\bibitem{chow2022deeprest}
K.-H. Chow, U.~Deshpande, S.~Seshadri, and L.~Liu, ``Deeprest: deep resource estimation for interactive microservices,'' in \emph{Proceedings of the Seventeenth European Conference on Computer Systems}, 2022, pp. 181--198.

\bibitem{touvron2022resmlp}
H.~Touvron, P.~Bojanowski, M.~Caron, M.~Cord, A.~El-Nouby, E.~Grave, G.~Izacard, A.~Joulin, G.~Synnaeve, J.~Verbeek \emph{et~al.}, ``Resmlp: Feedforward networks for image classification with data-efficient training,'' \emph{IEEE transactions on pattern analysis and machine intelligence}, vol.~45, no.~4, pp. 5314--5321, 2022.

\end{thebibliography}

\clearpage
\newpage

\appendix
\subsection{Dataset}
\label{appendix:dataset}
\subsubsection{Data Collection}
\label{appendix:dataset-collection}

A core prerequisite for this study is the collection of a high-fidelity and well-synchronized dataset that captures both workload dynamics and resource utilization. However, acquiring such data represents a significant practical challenge, because many cloud platforms lack the native tooling required to monitor application-level traffic with sufficient granularity. This deficiency causes many prior studies~\cite{park2021graf, somashekar2024gamma} to neglect the critical impact of service interaction patterns on system performance. To overcome this limitation, we design and implement a robust monitoring pipeline centered around Istio~\cite{istio2025} and Prometheus~\cite{prometheus2025}. The proposed pipeline utilizes the data plane of Istio, which intercepts all network traffic through the Envoy Sidecar proxies deployed alongside each microservice pod. This interception facilitates the collection of detailed traffic stream telemetry, such as request volumes and response latencies. Concurrently, a cluster-wide Prometheus instance scrapes resource stream metrics, such as the utilization of CPU and memory, from each node and pod. The critical integration point of the proposed approach is the configuration of Prometheus to simultaneously scrape the telemetry endpoints exposed by the Envoy proxies. This configuration unifies the heterogeneous data streams into the same Prometheus Time-Series Database (TSDB). This design ensures precise temporal synchronization and provides a solid foundation for the subsequent feature engineering. Table \ref{tab:features} provides an exhaustive list of the specific metrics collected through this monitoring pipeline.

\subsubsection{Dataset Characteristics}
\label{appendix:dataset-characteristics}

\begin{figure}[!ht]
\centering
\includegraphics[width=0.95\columnwidth]{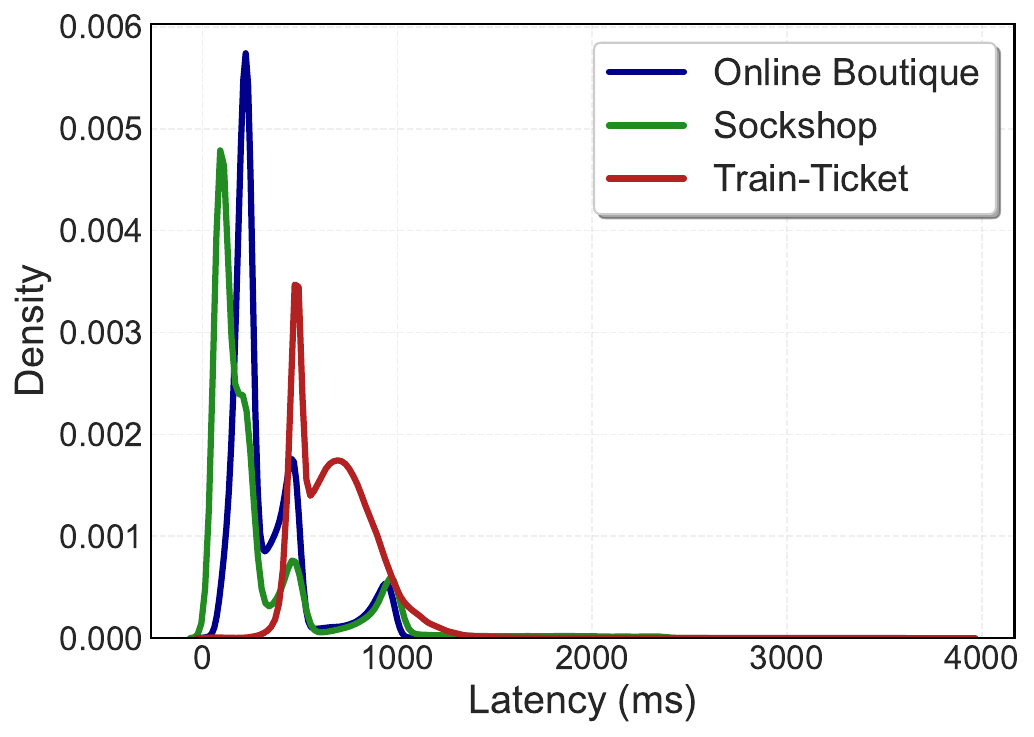}
\caption{Distribution of P95 latency across the evaluated benchmarks. The distinct peaks suggest different system operational modes, while the long tail represents rare, high-latency events.}
\label{distribution}
\end{figure}

A detailed analysis of the window-level P95 latency reveals a complex data distribution, which complicates the prediction task. As illustrated in Fig.~\ref{distribution}, the latency distributions across all three benchmarks exhibit a right-skewed, long-tail shape and, more notably, a multi-peak characteristic. The presence of several distinct peaks, rather than a single smooth decay, suggests that the systems operate in discrete performance regimes that correspond to normal, congested, and saturated states. Notably, the Train Ticket benchmark operates at a higher baseline latency compared to the e-commerce systems. Despite this difference in scale, the distribution maintains the fundamental multi-peak and long-tail characteristics observed in the e-commerce benchmarks. This multi-state behavior presents a modeling challenge that simple regression models cannot adequately address.

The statistical data presented in Table~\ref{tab:latency_stats_transposed} quantifies these visual observations. Across the datasets, high standard deviations relative to the respective means indicate extreme performance volatility. The right-skewed nature is evidenced by the disparity between the mean and the median for the Online Boutique~\cite{onlineboutique2025} and Sock Shop~\cite{sockshop2025} benchmarks. Although the Train Ticket benchmark~\cite{zhou2018benchmarking} displays a closer mean and median because of a higher sustained operational load, the maximum latency reaches nearly 4000 ms. This observation confirms the presence of extreme outliers. This long-tail distribution demonstrates how a minority of high-latency events disproportionately inflates the average response times of the system.

These empirical characteristics impose specific requirements on effective modeling approaches. The multi-peak nature implies the existence of distinct system-wide operational states, which requires models to possess sufficient expressive power to learn state-aware representations capable of distinguishing between different performance regimes. Furthermore, the long-tail distribution renders conventional loss functions, such as Mean Squared Error, highly susceptible to the influence of extreme values. This vulnerability necessitates a loss function that remains robust to outliers and invariant to the scale of the target variable.

The architectural and training decisions in this study directly respond to these data-driven requirements. The design of USRFNet captures the complex dynamics of the different performance regimes identified in the multi-peak distributions by learning a unified system representation from heterogeneous metrics. We adopt the Asymmetric Percentage Huber Loss~\cite{park2021graf} to mitigate the impact of the observed long-tail behavior, which ensures stable training and reliable predictions even during extreme latency events.

\begin{table*}[htbp]
\caption{Features Collected from Monitoring Pipeline}
\centering
\small
\resizebox{\textwidth}{!}{
\begin{tabular}{@{}llcl@{}}
\toprule
\textbf{Category} & \textbf{Metric Name} & \textbf{Target} & \textbf{Description / Query Hint} \\ \midrule
\multirow{5}{*}{resource stream} & container\_cpu\_usage\_seconds\_total & Node & Cumulative CPU time consumed by a service pod. \\
 & container\_memory\_usage\_bytes & Node & Memory usage of a service pod. \\
 & container\_spec\_cpu\_period & Node & CPU period allocated to the container. \\
 & container\_network\_receive\_bytes\_total & Node & Cumulative network bytes received by a pod. \\
 & container\_network\_transmit\_bytes\_total & Node & Cumulative network bytes transmitted by a pod. \\ 

\midrule

\multirow{6}{*}{traffic stream} & \multirow{2}{*}{istio\_requests\_total} & Node & Total requests received by a service (query by \texttt{destination\_workload}). \\
 &  & Edge & Requests from a source to a destination service (by \texttt{source\_workload, destination\_workload}). \\ \cmidrule(l){2-4}
 & \multirow{2}{*}{istio\_request\_bytes\_sum} & Node & Total request bytes received by a service (query by \texttt{destination\_workload}). \\
 &  & Edge & Request bytes from a source to a destination service (by \texttt{source\_workload, destination\_workload}). \\ \cmidrule(l){2-4}
 & \multirow{2}{*}{istio\_response\_bytes\_sum} & Node & Total response bytes sent from a service (query by \texttt{source\_workload}). \\
 &  & Edge & Response bytes from a source to a destination service (by \texttt{source\_workload, destination\_workload}). \\ \bottomrule
\end{tabular}%
}

\label{tab:features}
\end{table*}

\begin{table}[htbp]
\caption{Descriptive Statistics of the Target P95 Latency}
\centering
\small
\begin{tabular}{@{}cccc@{}}
\toprule
\textbf{Statistic}      & \textbf{Online Boutique} & \textbf{Sock Shop} & \textbf{Train Ticket} \\
\midrule
Count           & 69,121    & 71,696    & 85,209 \\
Min (ms)        & 24.13     & 46.25     & 34.06 \\
Max (ms)        & 2335.74   & 3675.51   & 3904.41 \\
Mean (ms)       & 338.99    & 298.60    & 679.84 \\
Std. Dev. (ms)  & 231.67    & 344.97    & 204.25 \\
25\% (Q1) (ms)  & 204.90    & 93.03     & 496.76 \\
Median (Q2) (ms)& 240.71    & 168.29    & 656.73 \\
75\% (Q3) (ms)  & 427.58    & 325.93    & 805.56 \\
\bottomrule
\end{tabular}
\label{tab:latency_stats_transposed}
\end{table}

\subsection{Experiments Hyperparameter Settings}
\label{appendix:experiments}

Table~\ref{tab:hyperparameters} details the hyperparameter configurations used to train the final USRFNet model. We selected these settings based on the optimal validation performance for each benchmark. To demonstrate the model's general robustness, core parameters governing the training process and the asymmetric loss function remained consistent across all three datasets. However, we adapted specific architectural parameters, such as the depth of the GNN~\cite{shi2021masked} and gMLP~\cite{liu2021pay} encoders, to accommodate the varying topological complexities of each application. Additionally, the RAGM modulation coefficients ($\alpha$ and $\beta$) were tailored to address the distinct optimization imbalances inherent to each system's scale.

\begin{table}[htbp]
\caption{Hyperparameter settings for USRFNet across the three benchmark datasets.}
\centering
\small
\setlength{\tabcolsep}{3pt} 
\begin{tabular}{@{}lccc@{}}
\toprule
\textbf{Hyperparameter} & 
\begin{tabular}{@{}c@{}}\textbf{Online} \\ \textbf{Boutique}\end{tabular} & 
\begin{tabular}{@{}c@{}}\textbf{Sock} \\ \textbf{Shop}\end{tabular} & 
\begin{tabular}{@{}c@{}}\textbf{Train~} \\ \textbf{Ticket}\end{tabular} \\
\midrule
Learning Rate & 1e-3 & 1e-3 & 1e-3 \\
Batch Size & 32 & 32 & 32 \\
Dropout (traffic stream) & 0.1 & 0.1 & 0.1 \\
Dropout (resource stream) & 0.1 & 0.1 & 0.1 \\
Epochs & 500 & 500 & 500 \\
\midrule
Traffic Node Features ($d_{n}$) & 3 & 3 & 3 \\
Traffic Edge Features ($d_{e}$) & 3 & 3 & 3 \\
Resource Features ($d_{r}$) & 5 & 5 & 5 \\
Embedding Dimension ($d_{emb}$) & 16 & 16 & 16 \\
GNN Layers (traffic stream) & 4 & 4 & 4 \\
gMLP Blocks (resource stream) & 4 & 4 & 5 \\
HIDAC Fusion Rank ($k$) & 4 & 4 & 4 \\
\midrule
Suppression Coefficient ($\alpha$) & 0.7 & 0.1 & 0.7 \\
Encouragement Coefficient ($\beta$) & 0.5 & 0.3 & 0.9 \\
\midrule
\multicolumn{4}{l}{\textit{Asymmetric Loss Parameters}} \\
Under-prediction Slope ($\alpha_L$) & 8.0 & 8.0 & 8.0 \\
Over-prediction Slope ($\alpha_R$) & 4.0 & 4.0 & 4.0 \\
Huber Thresholds ($\theta_L, \theta_R$) & 0.2 & 0.2 & 0.2 \\
\bottomrule
\end{tabular}

\label{tab:hyperparameters}
\end{table}

The variation in RAGM coefficients across datasets reflects the inherently different degrees of optimization imbalance in each system; as shown in Appendix~\ref{sec:sensitivity}, the overall MAPE variance across the full parameter grid remains within $0.5\%$, confirming that USRFNet does not rely on precise hyperparameter tuning to outperform baselines.

\subsubsection{Additional Baseline Comparison}
\label{appendix:baseline_comp}
The performance of USRFNet is evaluated against a comprehensive set of predictive models, as detailed in Table \ref{table:appendix_overall}. These baselines encompass traditional machine learning methods, standard neural networks, advanced tabular deep learning architectures, and advanced graph-based architectures.

\begin{table*}[ht]
\caption{Overall performance evaluation on three benchmarks.}
\centering
\setlength{\tabcolsep}{3pt} 
\begin{tabularx}{\textwidth}{c | *{3}{>{\centering\arraybackslash}X} | *{3}{>{\centering\arraybackslash}X} | *{3}{>{\centering\arraybackslash}X}}
\toprule
\multirow{2}{*}{\textbf{Model}} & \multicolumn{3}{c|}{\textbf{Online Boutique}} & \multicolumn{3}{c|}{\textbf{Sock Shop}} & \multicolumn{3}{c}{\textbf{Train Ticket}} \\ 
\cmidrule{2-10}
 & MAPE(\%) & MAE(s) & RMSE(s) 
 & MAPE(\%) & MAE(s) & RMSE(s) 
 & MAPE(\%) & MAE(s) & RMSE(s) \\
\midrule
GBDT   & 25.16 & 0.077 & 0.117 & 13.36 & 0.037 & 0.077 & 17.18 & 0.107 & 0.139 \\
MLP    & 16.63 & 0.049 & \underline{0.077} & 10.96 & \underline{0.031} & 0.075 & 14.08 & 0.090 & 0.126 \\
Transformer   & 11.92 & 0.042 & 0.090 & 9.59 & 0.039 & 0.125 & 12.24 & 0.086 & 0.129 \\
ResNet-like   & 12.06 & 0.043 & 0.099 & 9.88 & 0.037 & 0.085 & 10.81 & 0.079 & 0.123 \\
FT-Transformer   & 12.38 & 0.041 & 0.086 & 9.95 & 0.040 & 0.124 & \underline{9.22} & \underline{0.064} & 0.105 \\
GRAF   & 14.82 & 0.056 & 0.102 & 9.75  & 0.042 & 0.096 & 12.60 & 0.099 & 0.148 \\
TransformerConv   & 16.69 & 0.062 & 0.109 & 9.89  & 0.039 & 0.092 & 13.10 & 0.092 & 0.130 \\
GCN\textsuperscript{+} & 12.19 & 0.041 & 0.081 & 9.18 & 0.033 & 0.076 & 10.78 & 0.072 & 0.106 \\
GIN\textsuperscript{+} & 12.37 & \underline{0.040} & 0.080 & 9.48 & 0.034 & 0.076 & 10.19 & 0.068 & 0.103 \\
GatedGCN\textsuperscript{+} & \underline{11.91} & 0.042 & 0.084 & \underline{9.17} & 0.032 & \underline{0.073} & 9.86 & 0.067 & \underline{0.098} \\
\rowcolor{gray!20} \textbf{USRFNet} & \textbf{8.80} & \textbf{0.030} & \textbf{0.070} & \textbf{7.41} & \textbf{0.026} & \textbf{0.064} & \textbf{7.78} & \textbf{0.054} & \textbf{0.084} \\
\bottomrule
\end{tabularx}
\label{table:appendix_overall}
\end{table*}

The empirical results demonstrate that USRFNet consistently maintains a significant lead over all compared methods across the evaluated datasets. Specifically, the introduction of the ResNet-like~\cite{touvron2022resmlp} and FT-Transformer~\cite{gorishniy2021revisiting} models evaluates the efficacy of advanced deep learning architectures designed for tabular data. Although the FT-Transformer model achieves competitive results under specific workload conditions, the performance of this model remains less robust across diverse application scales when compared to advanced graph-based baselines. Furthermore, both tabular models process system features as independent vectors, which inherently neglects the topological dependencies among microservices. Consequently, the overall performance of these models remains inferior to the performance of USRFNet.

Notably, the performance of TransformerConv as a standalone model is significantly inferior to the performance of the complete USRFNet framework. Across the evaluated systems, the prediction error of the standalone graph convolution remains substantially higher than the error generated by the proposed framework. This performance gap confirms that although transformer-based graph convolutions excel at modeling workload propagation through service dependencies, these convolutions cannot achieve high precision without integrating the saturation constraints of infrastructure resources and synchronizing the learning process through gradient modulation. The results further validate that USRFNet effectively addresses the semantic misalignment and optimization imbalance inherent in multi-source observability data.

\subsubsection{Additional Ablation Analysis}
\label{appendix:ablation}
To justify the necessity of each component, Table~\ref{table:appendix_ablation} presents an exhaustive ablation study involving 14 variants (C1--C14). Specifically, the variants are defined as follows. \textbf{C1}: traffic stream only (node and edge features) with the GNN encoder, discarding resource metrics. \textbf{C2}: resource stream only with the gMLP encoder, discarding traffic metrics. \textbf{C3}: gMLP resource encoder replaced by GATv2 over the service dependency graph. \textbf{C4}: gMLP resource encoder replaced by FT-Transformer. \textbf{C5}: traffic encoder's Transformer-based graph convolution replaced by GATv2. \textbf{C6}: traffic encoder replaced by GraphSAGE, ignoring edge attributes. \textbf{C7}: Low-Rank Tensor Fusion in HIDAC replaced by standard cross-attention only. \textbf{C8}: Cross-Diffusion-Attention stage removed from HIDAC, feeding raw embeddings directly into tensor fusion. \textbf{C9}: RAGM disabled, training with a naive summation of stream losses. \textbf{C10}: traffic encoder replaced by standard GCN with uniform neighborhood aggregation and no edge features. \textbf{C11}: gMLP resource encoder replaced by a ResNet-like MLP without spatial gating. \textbf{C12}: HIDAC replaced by simple concatenation of $\mathbf{z}_t$ and $\mathbf{z}_r$. \textbf{C13}: HIDAC replaced by gated fusion, applying a learned scalar gate before addition. \textbf{C14}: HIDAC replaced by direct element-wise addition of $\mathbf{z}_t$ and $\mathbf{z}_r$.

The evaluation begins with the verification of the dual-modality input through the traffic-only variant (C1) and the resource-only variant (C2). The results demonstrate that USRFNet consistently outperforms both variants. This observation confirms that neither software workload propagation nor localized infrastructure capacity can independently capture the holistic operational state of a microservice system. Furthermore, variant C3 replaces the gMLP encoder on the resource stream with a graph neural network (GATv2). By attempting to process resource metrics through explicit topological edges, this variant compromises the physical decoupling of the architecture and exhibits a substantial increase in prediction error. This performance degradation justifies the use of the proposed dual-stream design to prevent semantic misalignment between propagation-based traffic signals and saturation-based resource signals.

\begin{table*}[ht]
\centering
\caption{Evaluation results of the exhaustive ablation study. Variants C1--C14 justify the necessity of dual-stream inputs, sub-network architectures, fusion strategies, and core innovation modules.}
\label{table:appendix_ablation}
\setlength{\tabcolsep}{3pt}
\begin{tabularx}{\textwidth}{c | *{3}{>{\centering\arraybackslash}X} | *{3}{>{\centering\arraybackslash}X} | *{3}{>{\centering\arraybackslash}X}}
\toprule
\multirow{2}{*}{\textbf{Variant}} & \multicolumn{3}{c|}{\textbf{Online Boutique}} & \multicolumn{3}{c|}{\textbf{Sock Shop}} & \multicolumn{3}{c}{\textbf{Train Ticket}} \\
\cmidrule{2-10}
 & MAPE(\%) & MAE(s) & RMSE(s) & MAPE(\%) & MAE(s) & RMSE(s) & MAPE(\%) & MAE(s) & RMSE(s) \\
\midrule
C1 & 32.12 & 0.143 & 0.254 & 24.51 & 0.106 & 0.246 & 15.38 & 0.109 & 0.160 \\
C2 & 10.75 & 0.037 & 0.076 & 9.06 & 0.032 & 0.076 & 13.97 & 0.101 & 0.149 \\
C3 & 16.84 & 0.060 & 0.110 & 10.08 & 0.035 & 0.084 & 9.65 & 0.071 & 0.114 \\
C4 & 13.05 & 0.046 & 0.097 & 9.40  & 0.036 & 0.079 & 11.23 & 0.083 & 0.129 \\
C5 & 10.26 & 0.035 & \underline{0.071} & 8.72  & 0.030 & 0.073 & 12.66 & 0.092 & 0.139 \\
C6 & 9.75 & 0.034 & 0.080 & 8.61  & 0.032 & 0.082 & 10.47 & 0.072 & 0.109 \\
C7 & \underline{9.22} & 0.032 & 0.079 & 8.17  & \underline{0.028} & 0.073 & 10.30 & 0.070 & 0.106 \\
C8 & 11.31 & 0.039 & 0.080 & 8.65  & 0.030 & 0.072 & 12.39 & 0.091 & 0.138 \\
C9 & 9.31 & \underline{0.031} & 0.075 & \underline{8.05} & 0.029 & \underline{0.070} & \underline{8.32} & \underline{0.058} & \underline{0.092} \\
C10 & 9.84 & 0.035 & 0.083 & 8.22  & 0.029 & 0.076 & 12.08 & 0.086 & 0.128 \\
C11 & 13.12 & 0.047 & 0.098 & 9.56  & 0.035 & 0.077 & 9.66 & 0.064 & 0.104 \\
C12 & 9.74 & 0.034 & 0.078 & 8.36  & 0.031 & 0.078 & 12.46 & 0.088 & 0.130 \\
C13 & 9.85 & 0.034 & 0.079 & 8.13  & 0.030 & 0.077 & 11.57 & 0.081 & 0.121 \\
C14 & 9.67 & 0.033 & 0.076 & 9.42  & 0.035 & 0.088 & 9.70 & 0.068 & 0.109 \\
\rowcolor{gray!20}\textbf{USRFNet} & \textbf{8.80} & \textbf{0.030} & \textbf{0.070} & \textbf{7.41} & \textbf{0.026} & \textbf{0.064} & \textbf{7.78} & \textbf{0.054} & \textbf{0.084} \\
\bottomrule
\end{tabularx}
\end{table*}

\begin{figure*}[ht]
    \centering
    \begin{subfigure}[b]{0.32\textwidth}
        \centering
        \includegraphics[width=\linewidth]{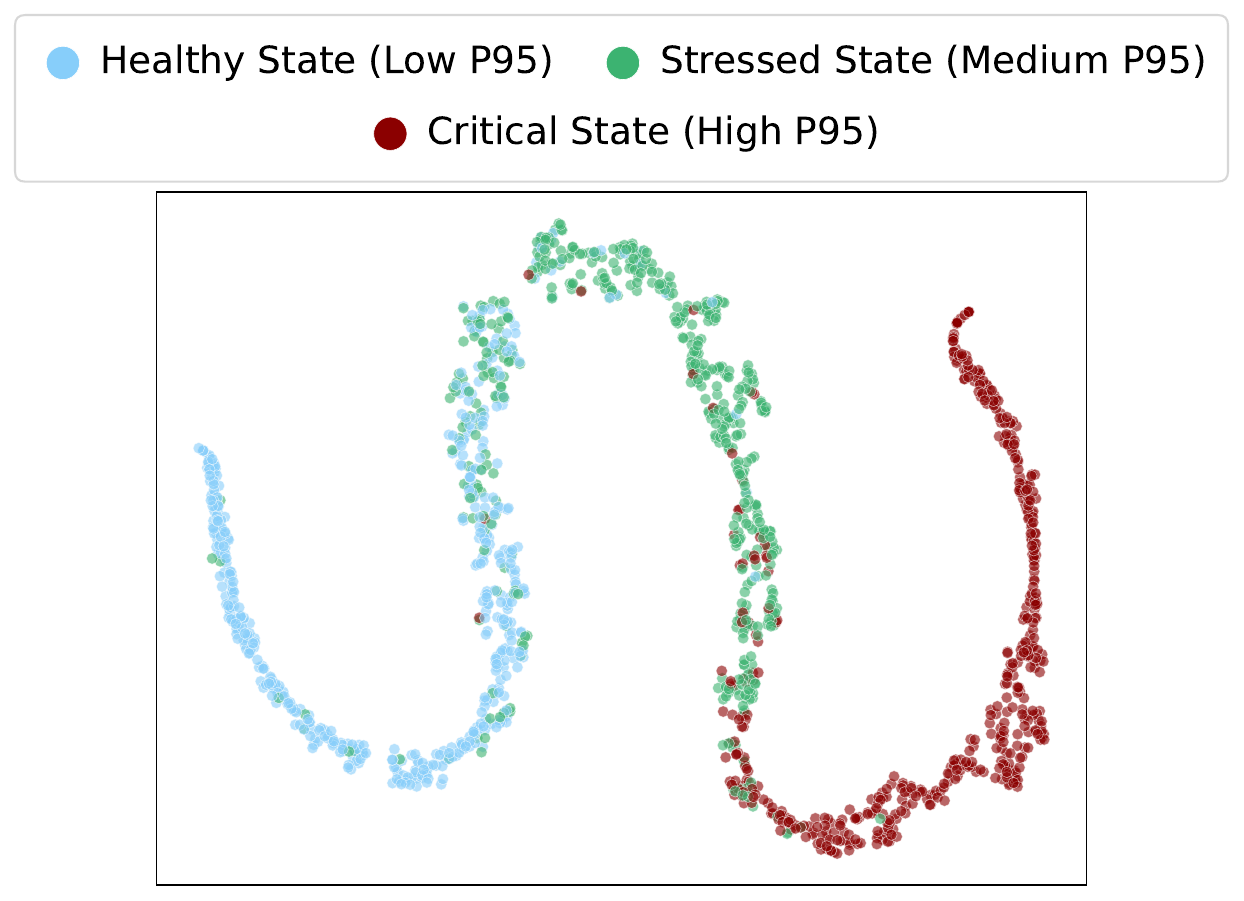}
        \caption{Online Boutique}
        \label{fig:tsne_boutique}
    \end{subfigure}
    \hfill
    \begin{subfigure}[b]{0.32\textwidth}
        \centering
        \includegraphics[width=\linewidth]{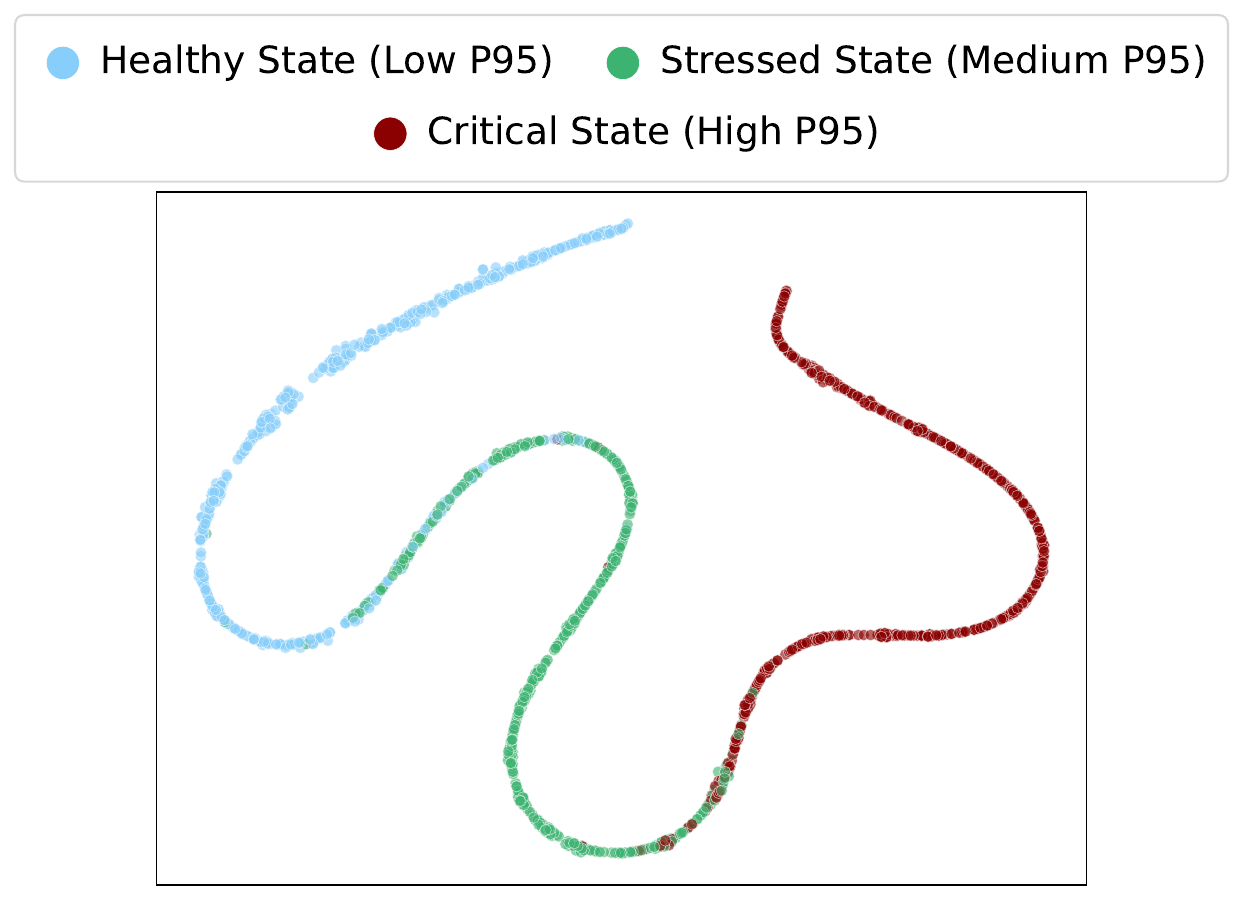}
        \caption{Sock Shop}
        \label{fig:tsne_sockshop}
    \end{subfigure}
    \hfill
    \begin{subfigure}[b]{0.32\textwidth}
        \centering
        \includegraphics[width=\linewidth]{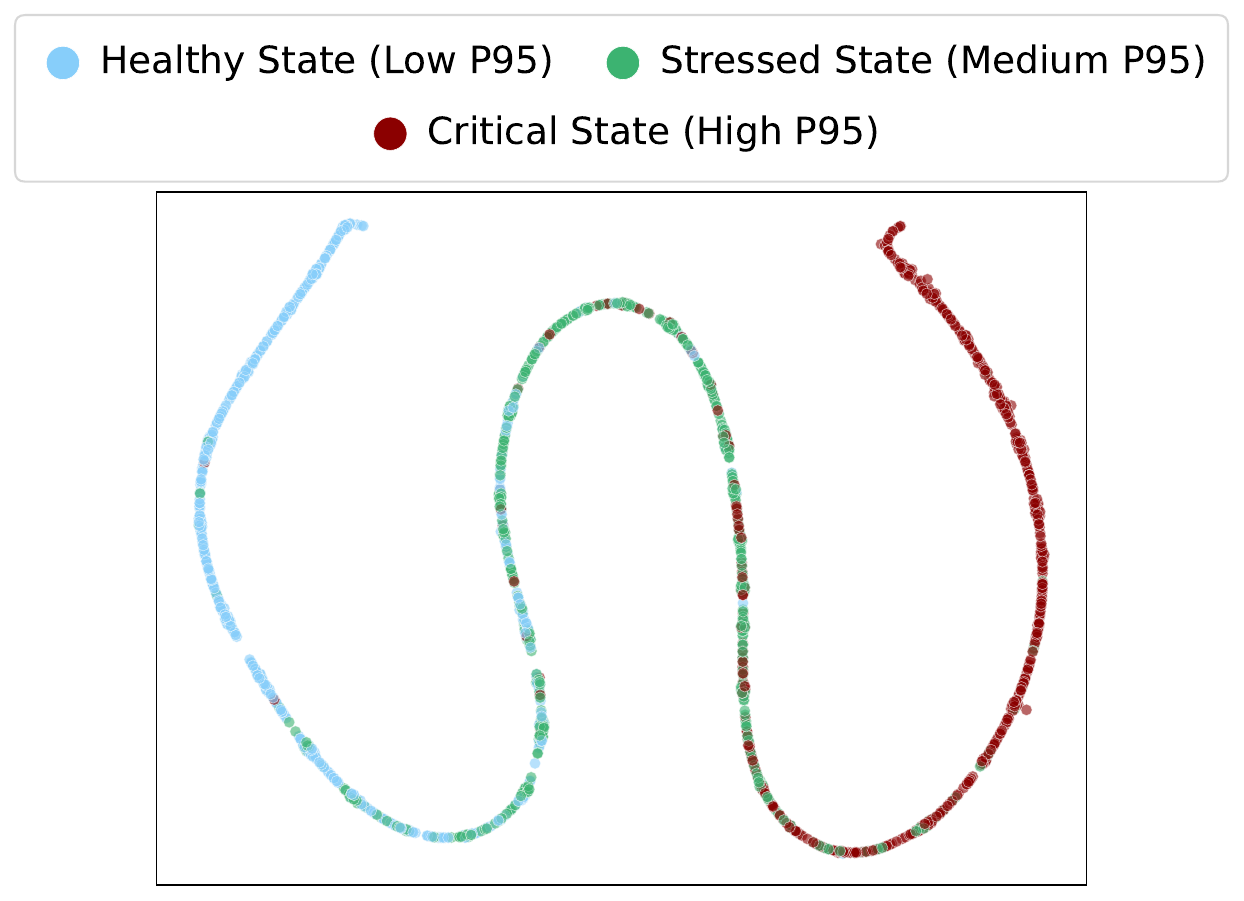}
        \caption{Train Ticket}
        \label{fig:tsne_trainticket}
    \end{subfigure}
    \caption{t-SNE visualization of the system embeddings generated by USRFNet across three benchmarks. The embeddings are color-coded by performance states, demonstrating clear clustering and discriminative power regarding different levels of tail latency.}
    \label{fig:tsne_all}
\end{figure*}

\begin{figure*}[!ht]
\centering
\includegraphics[width=1\textwidth]{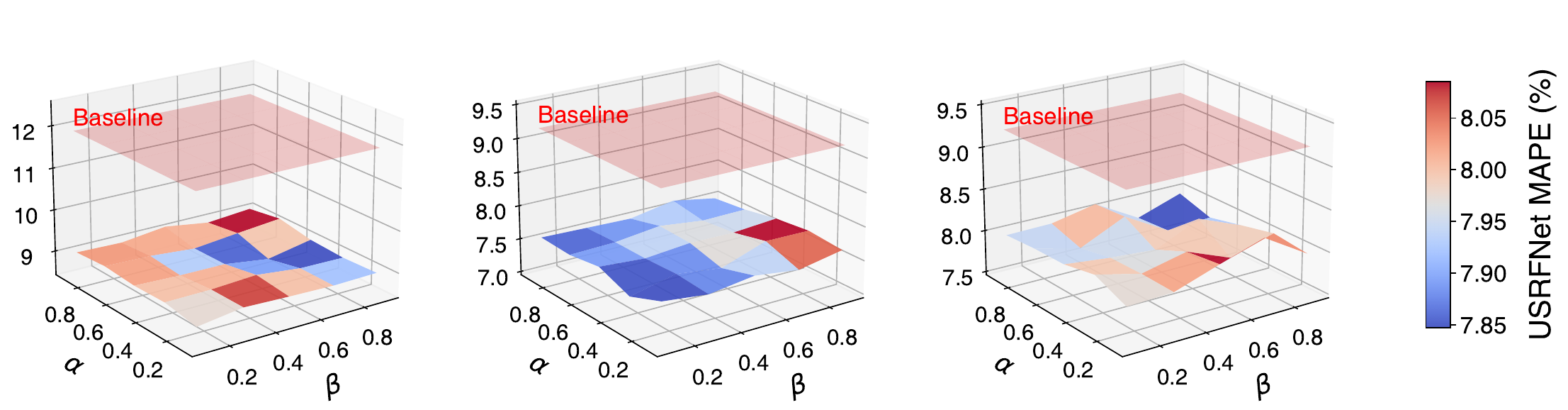}
\caption{Robustness evaluation of the RAGM mechanism. The 3D surfaces display the MAPE of USRFNet across varying combinations of the suppression ($\alpha$) and encouragement ($\beta$) coefficients. The semi-transparent red planes represent the MAPE of the best-performing baseline for each dataset. The relatively flat error surfaces indicate minimal performance variance, whereas the consistent positioning strictly below the baseline planes confirms that the worst-case performance of USRFNet remains superior to state-of-the-art models.}
\label{fig:sensitivity}
\end{figure*}

Variants C4--C6, C10, and C11 further validate the selection of the sub-networks within each stream. Within the traffic stream, the transformer-based graph convolution of USRFNet is compared with variants where the traffic encoder is replaced by GATv2 (C5), GCN (C10), or GraphSAGE (C6). The superiority of the proposed traffic encoder arises from the capability to map edge attributes directly into the attention keys. Unlike variants C10 or C6, which assign fixed weights or ignore the intensity of the edges, USRFNet prioritizes neighboring nodes based on the actual magnitude of the traffic. This prioritization remains essential for modeling the dynamic fluidity of system demand. Within the resource stream, the gMLP-based encoder is compared with the replacement of the resource sub-network with an FT-Transformer (C4) or a ResNet-like architecture (C11). The gating-based design achieves higher stability by capturing implicit global contention through spatial gating. This characteristic remains frequently neglected by standard attention-based tabular models or residual networks.

The evaluation of integration strategies clarifies why traditional fusion methods are inferior to the HIDAC module. Simple concatenation (C12) treats demand and capacity features as independent vectors, which fails to characterize the complex non-linear coupling between the workload and the resources. Although cross-attention (C7) provides semantic alignment, the associated dot-product mechanism is primarily linear in nature, which makes this approach insufficient for modeling the multiplicative interactions, such as queuing effects, that dictate tail latency. Gated fusion (C13) merely rescales individual features and lacks the higher-order interaction modeling required to represent critical system bottlenecks. Direct addition (C14) induces detrimental signal interference by forcing divergent semantic spaces into a unified magnitude. In contrast, the HIDAC module utilizes Cross-Diffusion-Attention for contextual calibration and Low-Rank Tensor Fusion for multiplicative interaction, which provides a more expressive representation. Finally, variants C8 and C9 justify the core innovation modules. The variant without Cross-Diffusion-Attention (C8) demonstrates that semantic misalignment hinders the identification of demand-capacity bottlenecks. The variant without the RAGM mechanism (C9) proves that gradient modulation remains indispensable for mitigating optimization discrepancies. Without RAGM, the model overfits to simpler resource patterns and neglects the structural insights embedded within the service graph. These results confirm that every component contributes critically to the predictive accuracy of USRFNet.

\subsubsection{Visualization of System Embeddings}
To qualitatively evaluate the representational capacity of the system embeddings, this study employs t-Distributed Stochastic Neighbor Embedding (t-SNE) to project the high-dimensional latent vectors into a two-dimensional space. Fig. \ref{fig:tsne_all} illustrates the visualization results for the Online Boutique, Sock Shop, and Train Ticket benchmarks. The data points are categorized into three distinct performance states according to the observed window-level P95 latency: Healthy State (Low P95), Stressed State (Medium P95), and Critical State (High P95).

As depicted in the visualization, USRFNet demonstrates a robust capability to cluster the embeddings based on the underlying performance of the system. Across all three benchmarks, the data points that belong to the same performance category form well-defined clusters with clear decision boundaries. For example, the critical states characterized by high P95 latency remain significantly separated from the healthy states. This separation suggests that the decoupled dual-stream architecture of USRFNet effectively captures the discriminative features of system degradation. Furthermore, the tight clustering within each state and the spatial separation between the different states confirm that the system embeddings are highly expressive. These embeddings provide a meaningful representation of the holistic operational status. Such interpretability remains essential for downstream tasks, which include root cause localization and proactive resource orchestration.

\subsubsection{Robustness of the RAGM Mechanism}
\label{sec:sensitivity}

We evaluate the structural robustness of the Reliability-Aware Gradient Modulation (RAGM) mechanism across different hyperparameter configurations. Because microservice environments vary significantly in scale and complexity, a practical gradient regulation strategy must remain effective across a broad parameter space without requiring exhaustive tuning. We conduct a grid search over the suppression coefficient $\alpha$ and the encouragement coefficient $\beta$ within the range of $[0.1, 0.9]$. As illustrated in Fig. \ref{fig:sensitivity}, we plot the resulting MAPE response surfaces against the performance threshold of the best-performing baseline for each dataset, which is represented by the semi-transparent red plane.

The 3D visualizations reveal that USRFNet maintains a significant performance advantage regardless of the exact parameter configurations across all evaluated benchmarks. The generated error surfaces exhibit minimal fluctuations, with the maximum MAPE variance constrained to approximately $0.5\%$ across the different datasets. More importantly, even under the most unfavorable parameter configurations, the entire error surfaces of USRFNet remain strictly below the respective baseline planes. This observation confirms that the superiority of the proposed dual-stream architecture does not rely on strict parameter fine-tuning. Instead, the RAGM mechanism provides a highly stable optimization landscape that consistently mitigates training imbalances. As long as explicit gradient modulation is applied, the model guarantees a lower-bound performance that substantially surpasses the capabilities of existing single-stream methods.

\subsection{Limitations}
\label{sec:limitations}

Although USRFNet demonstrates high efficiency, three limitations remain. First, the framework prioritizes instantaneous spatial states to ensure millisecond-level responsiveness, which inherently limits the exploitation of long-term temporal dependencies in telemetry data. While snapshot-based modeling minimizes computational overhead, further mining of historical trends could refine the predictive accuracy. Future work may investigate the integration of lightweight sequence modeling to capture temporal evolution without compromising the real-time constraints of the model. Second, although the system predicts end-to-end tail latency, it lacks a fine-grained mechanism to decompose latency contributions across individual microservices. Attributing latency fluctuations to specific service nodes remains critical for pinpointing intermediate bottlenecks in production environments. Third, the current formulation assumes a constant set of microservices $\mathcal{V}$ within the aggregate dependency graph. This assumption omits the dynamic registration of new services. However, this design choice relies on the principle of timescale isolation. The deployment of new service architectures typically aligns with software release cycles that span days, whereas USRFNet functions as a predictive engine for short-term autonomic control loops. Within such micro-scale operational windows, treating $\mathcal{V}$ as a constant provides a statistically sound approximation that maintains the inference efficiency of the model. Future research will address these limitations by exploring path-aware embeddings and dynamic graph expansion techniques to capture architectural evolutions and fine-grained service metrics.

\end{document}